\pdfoutput=1
\documentclass{article}
\usepackage[preprint]{neurips_2025}
\usepackage{hyperref}
\usepackage{url}
\usepackage{booktabs}
\usepackage{tabularx}
\usepackage{amsfonts}
\usepackage{amsmath}
\usepackage{nicefrac}
\usepackage{microtype}
\usepackage{graphicx}
\usepackage{natbib}
\usepackage{algorithm}
\usepackage{algorithmic}
\usepackage{adjustbox}
\usepackage[utf8]{inputenc}
\usepackage[T1]{fontenc}
\usepackage{lmodern}

\title{Wrong Design Intent Is Worse Than Never Conditioning: A Derangement-Control Diagnosis of Header Conditioning in CAD Program Completion}

\author{Yang Xiao \\ Sichuan University \\ \texttt{yangx64519@gmail.com}}


\begin{document}

\maketitle

\begin{abstract}
Design-intent conditioning of fine-tuned code LLMs is routine, but its correctness axis --- what happens when the intent is wrong rather than merely absent --- is untested, and its benefit is usually scored with the same feature detector that defines the conditioning signal. CADCON prepends a five-feature design-intent header to CadQuery-style sketch-extrude programs during LoRA fine-tuning of Qwen2.5-Coder-1.5B; adherence is scored by executable geometric assertions on the produced B-rep solid that share no code with the header-defining extractor. On 400 deduplicated held-out programs stratified over eleven four-tag intent profiles, at 40\% prefix, a pre-registered replication finds that a semantically wrong header degrades adherence below the never-header-trained baseline --- significantly on 3/3 seeds under both tokenizations at the program level, and on 3/3 token and 2/3 text seeds at the level of the 298 distinct model inputs those programs present, which is the unit we state results at. Wrong-header executability is not depressed relative to that baseline for the text form (0.638 against 0.603), so there the deficit is not a by-product of programs that fail to run; for the token form it is lower and an executability contribution is not excluded. A derangement control, retrained so that every program receives another program's header, holds the header marginal byte-identical while destroying the correlation, and saw the same 400 programs with byte-identical wrong headers at the same indices. Its correct-to-wrong adherence change is indistinguishable from zero and inconsistent in sign, while the standard model drops substantially; the per-sample interaction (text headers) is positive and significant on 3/3 seeds ($p \leq 5.9{\times}10^{-7}$). The model's sensitivity to whether the header is right or wrong therefore requires the learned header-to-program mapping and is not a mechanical reaction to an off-distribution input. The control clears its pre-registered competence gate but is weaker than the protocol anticipated; on executable outputs it sits near the unconditioned baseline rather than at a floor, so it has adherence to lose and does not lose it. Its shortfall against that baseline is the same under a correct header as under a wrong one (-0.063 against -0.066), so this design does not test whether the below-baseline level itself requires the mapping, and we claim the sensitivity and not the level. What the mapping demonstrably does is route content into geometry: on features the true intent does not contain, the standard model realizes the feature far more often when the wrong header names it than when it does not, on three of the four scored features and under both tokenizations, while the control shows no movement the data can detect. Under the independent metric the correct header's benefit roughly halves for the text form; for the token form the arm ordering is unit-dependent and reflects how often that form writes a program which runs (0.477 against 0.603) rather than whether it follows the header when it does. Ground truth itself scores only 0.567 on its own target under this metric --- not a ceiling, since the arms complete from a prefix and often run where the reference does not, but a statement of what the target is: agreement with a regex-declared attribute set rather than the fraction of a design intent achieved. Effects are strongly heterogeneous across intent families and per-profile estimates are descriptive, though the aggregate is not an artifact of the stratified draw. The effect belongs to conditional completion: without a program prefix the unconditioned baseline cannot generate valid CAD, leaving the contrast undefined. Decision rules, thresholds and analysis scripts were frozen before the corresponding data existed; a blinded power check preceded any computation of direction or p-value. Wrong design intent is not inert: it actively misdirects generation.
\end{abstract}

\section{Introduction}

\label{sec:introduction}

Parametric CAD encodes not just geometry but design intent --- the shape classes, proportions, and topology a part is meant to have. As code-pretrained LLMs are fine-tuned to emit sketch-extrude programs, systems increasingly attach a field that carries that intent into generation: a text description, a structured header, or an upstream model's inferred plan. Once a model is trained with such a field, the field is part of its input contract, and a question follows that is about deployment rather than about accuracy: what does the model do when the field's content is wrong? Upstream error is ordinary --- a stale specification, a mis-retrieved description, a plan inferred by another model --- and the comfortable assumption is that a bad signal at worst wastes the conditioning, leaving the model to degrade toward the behavior it would have shown unconditioned. That assumption has not been tested, and this paper reports that in our setting it is false.

Prior work conditions CAD generation on natural language, images, or point clouds and reports that conditioning ``works'' \cite{xie2025texttocadquery, guan2025cadcoder, wu2021deepcad}. Three questions have gone unexamined in structured geometric generation. First, what happens when the conditioning content is \textit{wrong}: does the model ignore it (graceful), or trust it and generate something worse than it would have unconditioned (hazardous)? Second, if wrong content hurts, is that because the model \textit{reads} the content --- or merely because any mismatched header is an out-of-distribution input for a model trained to expect a consistent one? Third, is the reported \textit{benefit} of conditioning real, or partly an artifact of scoring adherence with the same feature detector used to build the conditioning target?

We answer all three with CADCON (\textbf{CAD CON}ditioning), a five-feature design-intent header (CIRCLE, NGON, THIN, TALL, MULTI\_PART) prepended to CadQuery-style sketch-extrude programs --- transpiled from DeepCAD histories \cite{wu2021deepcad} --- during LoRA fine-tuning of Qwen2.5-Coder-1.5B \cite{hui2024qwen25coder}. Every claim this paper makes about wrong versus absent intent, and every number that carries one, is measured on a single frozen sample: 400 deduplicated held-out programs drawn stratified over all eleven four-tag intent profiles the clean held-out pool contains, at a 40\% program prefix, across three seeds (§3.2); material retained from the earlier evaluation --- the 0\%-prefix regime, the dose--response probe, and the studies collected in Appendix D --- is labelled with its run wherever it appears. Two instruments make the answers interpretable (Figure 1). The first is produced-side metric independence: adherence is re-scored by executable assertions on the produced B-rep solid (in the style of CADTests \cite{mallis2026cadtests}), sharing no code with the training-side regex extractor that defines the headers, so that a conditioning benefit cannot be credited by the detector that created the signal being followed (§3.4). The second is a causal control: we retrain the model with \textit{deranged} ground-truth headers --- every training program receives another program's header, so the header marginal is byte-identical while the header$\leftrightarrow$program correlation is destroyed --- and test whether this control, which cannot have learned a header$\rightarrow$program mapping, is also hurt by wrong headers at evaluation (§3.5). Both the wrong-header contrast and the control's dissociation are measured on the same 400 programs, in the same arm, with byte-identical wrong headers injected at the same indices, so the inferential and the causal evidence are not separated by sample. Decision rules, thresholds and analysis scripts were frozen with SHA-256 hashes in the two links of this campaign's freeze chain (§A.1): the replication protocol, fixed before any of its data existed, and the control increment, fixed while the control's arm on this sample held no data at all, so the interaction was pre-registered while its value was unknowable. A blinded power check, verified by AST inspection to contain no mean, direction or Wilcoxon path, ran before any effect direction or p-value was computed (§3.6). Throughout, ``causally controlled'' refers specifically to this design: it isolates the learned header$\rightarrow$program mapping as the cause of the model's sensitivity to header content, excluding the marginal/mechanical distribution-shift confound; the finer mechanism --- graded semantic reading versus learned consistency-checking --- is explicitly out of scope (§5).

An earlier, smaller and unstratified evaluation --- 76 scored entries covering 38 unique programs across six of the eleven profiles --- first produced this ordering, at 3/3 seeds direction and, at the 38-program unit, significance on 1/3 seeds text-side against a retrospective power of 0.463, and is what the pre-registered replication was written to settle. The two samples differ in intent-profile composition, by construction, and in generation-truncation regime, so their absolute levels are reported separately throughout and are never pooled or placed on a common scale (§3.2, §6); this is a condition of the frozen protocol and not a claim that either evaluation sits systematically above the other, whose direction varies by arm.

Our contributions:

\begin{enumerate}
  \item Under a pre-registered protocol, at 40\% prefix on 400 deduplicated held-out programs stratified over eleven four-tag intent profiles, a semantically wrong header degrades execution-inclusive geometric adherence below the never-header-trained baseline on 3/3 seeds under both tokenizations, significant on all six of the resulting cells at the program level, with no reverse-direction violation and a passing profile-direction guardrail. Because the 400 programs present only 298 distinct model inputs, the headline is stated at that unit, where five of six cells survive, the pre-registered $\geq$2/3-seed rule is met on both tokenizations, and hierarchical bootstrap intervals exclude zero for both. That the deficit is not an executability effect rests on the text arm running \textit{more} often than the baseline, not on the executable-intersection subset, which conditions on a post-treatment variable and is reported as descriptive (§4.1).
  \item The derangement-trained control R --- identical recipe, identical header marginal, no learnable header$\leftrightarrow$program relation --- saw the same 400 programs in the same arm, with byte-identical wrong headers at the same indices. R clears its pre-registered competence gate yet shows essentially no correct$\rightarrow$wrong drop where the standard model drops substantially; the per-sample interaction is positive and significant on 3/3 seeds, survives removal of any single intent profile, and clears all three pre-registered guardrails --- non-zero-pair count, executable intersection and profile level. The model's \textit{content sensitivity} --- its response to whether the header is right or wrong --- therefore requires the learned header$\rightarrow$program mapping, and the marginal/mechanical distribution-shift confound is excluded. Routing is visible directly: on features the true intent does not contain, the standard model realizes a feature far more often when the wrong header names it than when it does not, under both tokenizations, while R shows no movement the data can detect. R sits below the unconditioned baseline by almost the same margin under a correct header as under a wrong one, so its own level is an offset of derangement training rather than a header effect; the design therefore does not test whether the below-baseline \textit{level} of the wrong arm requires the mapping, and we claim the sensitivity it does test. R is also weaker than the protocol anticipated and its gate carries only on the $\geq$2/3-seed rule, but a control with less to lose is answered by the executable-only view, where R holds an adherence near the unconditioned baseline and loses 0.007 of it while the standard model loses 0.326 (§4.2, §6). To our knowledge this is the first causally controlled isolation of wrong-content sensitivity in structured geometric program generation.
  \item For the wrong-vs-none contrast, per-profile point estimates are strongly heterogeneous across intent families, three of eleven run positive, and all are descriptive: the pre-registered analysis carries no per-profile test, p-value or power, and no claim is attached to an individual intent family (§4.3). The aggregate is nonetheless not an artifact of which profiles the stratified draw over-samples, since a report-only reweighting to natural held-out frequencies strengthens rather than weakens it (§4.1); the inferential population is profile-balanced.
  \item The effect belongs to conditional completion: with no program prefix the unconditioned baseline cannot generate valid CAD at all, so ``wrong < never conditioning'' is undefined there. That regime was not re-tested, because its effective sample size is capped by header combinatorics at nineteen distinct prompts however many programs are drawn (§4.5, §6).
  \item Execution-level assertions that share no code with the extractor defining the conditioning signal, paired with a derangement-trained control, form an audit pair for conditioning claims: together they answer whether a reported benefit survives an independent metric and whether a model's sensitivity to the content of a conditioning signal requires the learned mapping. Here the independent metric roughly halves the text header's correct-arm gain and leaves the token header's indistinguishable from the unconditioned baseline at the unit this paper reports at; almost all of that fall is the cost of requiring the program to run, not of changing the detector (§4.4). What a signal-derived metric mainly hides is therefore that the conditioned model often writes a program which does not run, especially for the token form; hence the evaluation instrument is presented as method (§3.4), not a result.
\end{enumerate}

Together these results revise the default risk model of intent conditioning in structured generation --- a model in which, as §2 documents, work on conditioned structured generation tests whether the signal is present or how good it is, but never what happens when it is wrong. On the benefit side, conditioning in our setting delivers less than its own self-referential evaluation reports (contribution 5): the text header's gain roughly halves under an execution-level metric, and the token header's does not survive at all, which argues that conditioning claims in this literature should be read against evaluation that shares no code with the signal whose adherence is being scored. On the harm side, the worst-case assumption that a bad signal at most returns a model to its unconditioned behavior fails (contribution 1) --- a failure that extends to structured geometric generation what misleading-context work documents in natural language \cite{tao2026nwcad}. The control then locates the cause (contribution 2): because R was trained on the identical header marginal, receives byte-identical wrong headers at evaluation, and is nonetheless unmoved by them, the model's response to the header's content cannot be a mechanical reaction to a rare or malformed input; it requires the very mapping that makes a correct header useful. (The control bounds this to the content sensitivity and not to the below-baseline level itself; §4.2.) That localization matters for repair as much as for diagnosis. It rules out input validation as a mitigation --- a wrong header is another program's genuine header, and passes every format, vocabulary and schema check the correct one passes --- and it places the intervention at the conditioning interface itself: a represented option to proceed unconditioned, trained in, or a decode-time backoff toward an unconditioned stream when the signal is suspect. And because the harm lives in conditional completion (contribution 4) --- editing and extending an existing design, precisely the workflow such models are being built for --- the hazard sits in the intended primary use case rather than in a corner case.

The contribution is diagnostic rather than architectural, but the diagnosis carries a design principle, a portable method, and an agenda. The design principle: a trained conditioning field is a \textit{trust interface} --- once a model has learned to rely on it, upstream errors (a stale user specification, a retrieved description, or, as in plan-conditioned pipelines (§2.2), another model's inferred plan) are not filtered out but routed into geometry, so graceful degradation toward unconditioned behavior must be designed in, not retrofitted. The method: the audit pair used here --- an execution-level metric sharing no code with the extractor that defines the conditioning signal, plus a derangement-trained control that separates harm requiring the learned mapping from a mechanical reaction to an off-distribution input --- is not specific to CAD. Wherever the conditioning signal can be shuffled at training time and adherence can be scored independently of that signal, the pair offers a template for interrogating a ``the model follows X'' claim in conditioned generation, at a cost of pure-inference evaluation plus one additional fine-tuning run per seed. The agenda: whether the deflation and the harm persist at larger scale and across program representations; which content-dependent reading holds --- graded semantic reading or learned consistency-checking --- the grain our dose-response probe could not resolve (§5); how to detect or absorb wrong intent at decode time; and what governs the heterogeneity the replication exposes. One descriptive pattern in that heterogeneity is where we would look first: among the four single-feature profiles, the two this checkpoint renders least well without a header are the two a correct header helps most (pure-TALL 0.156 $\rightarrow$ 0.978, pure-NGON 0.179 $\rightarrow$ 0.786), and each of the three largest interaction point estimates in the eleven-profile table involves one of those two features (+0.867 pure-TALL, +0.533 NGON+TALL, +0.513 pure-NGON), so conditioning appears to do most of its work, and to carry most of its exposure, where the unconditioned model is weakest. This design tests no individual profile and we attach no claim to one (§4.3); the observation is reported as a target for a study that could. CADCON is the vehicle: a produced-side metric-independent, causally controlled, pre-registered testbed for characterizing when intent conditioning helps, when it is over-credited, and when it actively harms.

\section{Related Work}

\label{sec:related_work}

\subsection{Generative models for parametric CAD sequences}

\label{sec:generative_models_for_parametric_cad_seq}

Neural parametric CAD modeling began with dedicated architectures over construction sequences. DeepCAD \cite{wu2021deepcad} introduced the first generative model for CAD operation sequences, together with the large-scale dataset of sketch-extrude histories from which our corpus derives. SkexGen \cite{xu2022skexgen} encodes topological, geometric, and extrusion variation into disentangled codebooks, giving users coarse control over which properties the generated sequences share. Hierarchical neural coding \cite{xu2023hnccad} extracts a three-level code tree from a CAD model and completes a partial model conditioned on it, which makes it the closest published work on \textit{controllable} CAD generation: a user edits the code tree and the generation follows, though the edited tree defines the new target rather than making a claim about an existing one. SolidGen \cite{jayaraman2022solidgen} bypasses construction sequences altogether, autoregressively predicting vertices, edges, and faces to synthesize boundary representations directly. A second wave fine-tunes pretrained code LLMs. CADmium \cite{govindarajan2025cadmium} pairs more than 170k CAD models with human-like textual descriptions and fine-tunes a code LLM for text-driven sequential design. FlexCAD \cite{zhang2024flexcad} renders a CAD model as structured text and fine-tunes an LLM into a single controllable generator across all construction hierarchies, from sketch-extrusion down to individual curves. CAD-Recode \cite{rukhovich2024cadrecode} represents sketch-extrude sequences as executable Python code and translates point clouds into CadQuery programs --- the program representation also used here. CAD-Llama \cite{li2025cadllama} is the closest published analogue to our own setup: hierarchical semantic descriptions are prepended to a structured parametric CAD program and the pair is used for adaptive pretraining and instruction tuning, with an ablation that removes the description. CADFusion \cite{wang2025cadfusion} alternates supervised sequence learning with a visual-feedback stage that rewards sequences whose renders are preferred, varying which training signal is used rather than what an inference-time condition says. Across both waves, these systems model the distribution of valid CAD models and are evaluated on generation or reconstruction quality --- including, for the controllable ones, whether a requested property \textit{can} be produced. Whether the model is \textit{faithful} to the semantic content of a conditioning signal, and what happens when that content is wrong, sits outside their evaluation scope; that question is ours.

\subsection{Conditioning quality vs. conditioning presence}

\label{sec:conditioning_quality_vs_conditioning_pre}

Prior intent- and text-conditioned CAD generation varies the \textit{quality} --- or at most the \textit{presence} --- of a conditioning signal, but never its correctness. ToolCAD \cite{gong2026toolcad} evaluates text-to-CAD across the abstract-to-expert prompt levels (L0--L3) inherited from Text2CAD \cite{khan2024text2cad}, with every condition receiving some correct prompt and no unconditioned control; those levels vary how much parametric detail a prompt carries, not whether what it says is true of the target. CAD-Llama's description ablation \cite{li2025cadllama} is the same presence axis as CAD-HLLM's, on a system whose conditioning field is prepended to a code-formatted program much as ours is. ProCAD / ``Clarify Before You Draw'' \cite{yuan2026procad} detects under-specified or internally conflicting CAD prompts and repairs them via clarification, rather than measuring the cost of following a wrong one. Closest to our setting, CAD-HLLM \cite{zuo2025cadhllm} fine-tunes a parameter-completion model conditioned on an inferred symbolic plan --- a \textit{trained} conditioning field, like our header --- and ablates the plan away (showing generation degrades without it), yet never evaluates a plan that is \textit{wrong}: the correctness axis of a conditioning signal that the model has learned to depend on remains untested. One line of work does put design intent at the centre and should be distinguished explicitly, since the phrase is ours as well. Casey et al. \cite{casey2025aligning} align constraint generation \textit{toward} design intent, using a constraint solver as the reward to raise the share of fully-constrained sketches; intent there is a property of the output, optimized for, and the input is ground-truth sketch geometry. There is accordingly no conditioning field whose content could be made false. We hold the model fixed and vary whether the intent handed \textit{to} it is correct, which is the axis none of this work varies. In the natural-language setting, misleading context is known to depress conditioned generation below the no-context baseline: Tao \& Agrawal (2026, Findings of ACL) \cite{tao2026nwcad} formalize a ``do-no-harm'' requirement for context-conditioned generation --- naming \textit{neutral regression} the case where a model overwrites a correct answer under an effectively non-informative context --- and empirically document that plausible-but-misleading \textit{distractor} contexts push standard with-context decoding well below the no-context baseline (e.g., 60.4 $\rightarrow$ 36.2 for Llama-3.1-8B), which their two-stream decoder mitigates by backing off to the no-context stream. In retrieval-augmented QA, MADAM-RAG \cite{wang2025madamrag} mitigates misinformation in retrieved evidence and notes that a closed-book baseline can sometimes beat naive RAG; MisBench \cite{peng2025misbench} measures preference for in-context misinformation under knowledge conflict in multiple-choice QA. These establish that a wrong conditioning signal can be worse than none in natural language; none isolates this in conditional structured geometric generation --- nor, to our knowledge, does any pair it with a trained control that separates harm requiring the learned mapping from a mechanical reaction to input mismatch, as our derangement control does (§3.5).\footnote{We do not cite AlphaCode's metadata ablation as a wrong-vs-none precedent: on inspection it establishes no such ordering --- its incorrect-solution tag hurts only relative to the correct tag, and at the largest sample budget the no-tag condition is in fact lowest.}

\textit{(We deliberately do not adopt the term ``neutral regression'' for our effect: it is defined for non-informative context, whereas our wrong header carries incorrect content --- the analog is the distractor case above.)}

\subsection{Code models under corrupted or misleading signals (contrast)}

\label{sec:code_models_under_corrupted_or_misleadin}

Two lines of work perturb the signals a code model receives. Waheed et al. \cite{waheed2025codeinduced} corrupt code \textit{fine-tuning data} and find reasoning far more vulnerable to structural than to semantic corruption; CodeCrash \cite{lam2025codecrash} injects false comments and misleading hints into input code at inference time and degrades code-\textit{reasoning} accuracy. Neither varies the correctness of a conditioning field the model was trained to depend on: one perturbs training data, the other a channel the model never learned to read. That semantic corruption is often tolerable in code reasoning sharpens the interest of conditional structured \textit{generation} through such a field, where a semantically wrong signal is actively harmful.

\section{Method}

\label{sec:method}

\begin{figure}[!tb]
\centering
\includegraphics[width=0.97\columnwidth]{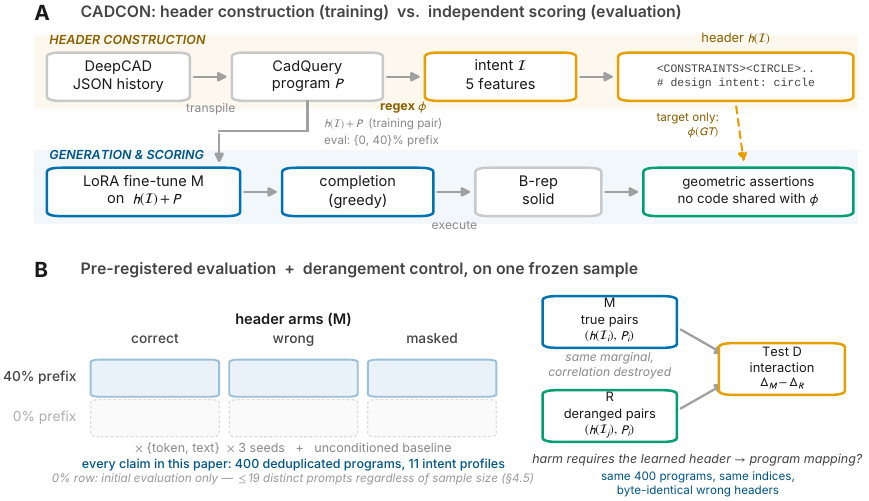}
\caption{\textbf{Method overview.} \textbf{(A)} Each DeepCAD history is transpiled to a CadQuery-style program; the regex extractor $\phi$ builds the design-intent header on the training side, while evaluation executes the completion and scores the produced solid with geometric assertions that share no code with $\phi$ (§3.4). \textbf{(B)} The pre-registered evaluation, left: the \{0\%, 40\%\}-prefix $\times$ \{correct, wrong, masked\} matrix over both tokenizations and three seeds, plus the never-header-trained unconditioned baseline (§3.6). Every claim in this paper is drawn from the 40\%-prefix row, re-run on the profile-stratified 400-program held-out sample (§3.2); the 0\% row was evaluated only in the initial evaluation, for the reason given in §4.5. Right: the derangement control --- R is trained on deranged header$\leftrightarrow$program pairs (identical header marginal, destroyed correlation) and Test D asks whether M's correct$\rightarrow$wrong adherence drop exceeds R's (§3.5).}
\label{fig:method_overview_a_each_deepcad}
\end{figure}

Figure 1 overviews the method: the header-construction pipeline and the independent scoring path \textbf{(A)}, and the pre-registered evaluation design including the derangement control \textbf{(B)}.

\subsection{Problem formulation and the design-intent header}

\label{sec:problem_formulation_and_the_design_inten}

Let a parametric CAD program $P$ be a sequence of sketch-extrude operations. Following the DeepCAD corpus convention \cite{wu2021deepcad}, each JSON construction history is transpiled by a deterministic script into a CadQuery-style sketch-extrude program --- a Python code string of \texttt{moveTo}/\texttt{lineTo}/\texttt{circle}/\texttt{extrude} calls. The model consumes the header-plus-program string through the base code LLM's own subword tokenizer, and $\mathbf{x}$ denotes that encoding. The transpiled language covers the core sketch-extrude subset --- straight-line and full-circle profiles with extrusions; arcs are approximated by straight segments and each sketch keeps its outer profile --- and the five features below are defined over this language.

Design intent is a sparse binary vector over five geometric properties --- CIRCLE (a circular profile), NGON (a polygonal loop), THIN and TALL (aspect-ratio thresholds) and MULTI\_PART (intended multi-solid geometry) --- extracted from the program text by a deterministic regex-based function requiring no human annotation:

\[
\mathcal{I} \;=\; \phi(P) \;\in\; \{0,1\}^5 ,
\qquad
\phi: P \mapsto \mathcal{I} .
\]

Writing $d$ for the extrude depth and $w$ for the largest planar extent of the profile, $\phi$ pattern-matches the program text as

\[
\begin{aligned}
\mathcal{I}_{\mathrm{CIRCLE}} &= 1 &&\iff && \text{the program calls a full-circle profile,}\\
\mathcal{I}_{\mathrm{NGON}} &= 1 &&\iff && \text{a closed profile loop has} \geq 5 \text{ straight segments,}\\
\mathcal{I}_{\mathrm{THIN}} &= 1 &&\iff && d/w < 0.3 ,\\
\mathcal{I}_{\mathrm{TALL}} &= 1 &&\iff && d/w > 2.0 ,\\
\mathcal{I}_{\mathrm{MULTI\_PART}} &= 1 &&\iff && \text{the program has more than one extrude block.}
\end{aligned}
\]

The last of these is an intent proxy that cannot itself see whether the extrudes fuse into one solid --- the under-determination §3.4 addresses.

CADCON prepends a header $h(\mathcal{I})$ to the program and fine-tunes with the standard causal cross-entropy loss over the concatenation, header tokens included:

\[
z \;=\; \bigl(h(\mathcal{I}),\, \mathbf{x}\bigr) ,
\qquad
\mathcal{L}(\theta) \;=\; -\sum_{t=1}^{|z|} \log p_\theta\!\left(z_t \mid z_{<t}\right) .
\]

The model thus learns

\[
p_\theta\!\left(\mathbf{x}_{k:} \mid h(\mathcal{I}),\, \mathbf{x}_{:k}\right) ,
\]

where the program prefix $\mathbf{x}_{:k}$ --- the first 40\% of the program text, a character-level strict prefix, empty in the 0\% regime --- is supplied only at evaluation: training always sees complete header-plus-program sequences. The same extractor $\phi$ defines both the training-time headers and the original regex adherence metric, the circularity that §3.4 breaks on the produced side.

\subsection{Data, representation, and the profile-stratified held-out sample}

\label{sec:data_representation_and_the_profile_stra}

The corpus derives from DeepCAD (\textasciitilde{}180,000 models as JSON sketch-extrude histories; in DeepCAD's native 256-token serialization these have mean length $\approx$150 and maximum 1,024 tokens). Each history is transpiled to a CadQuery-style program (§3.1); under the base model's tokenizer every training sequence fits complete --- EOS included --- within the 1,024-token cap of Table 2. Training uses 10,000 programs from the unmodified training partition. On the held-out side, exact- and loose-hash deduplication filters the published test split against train: of 15,848 candidate held-out programs, 6,406 are identified as train-leaked and removed, leaving 9,442 clean held-out programs. Two header tokenizations test whether conditioning behavior depends on surface form: \textbf{token} --- the active features as dedicated special tokens wrapped in delimiters (\texttt{<CONSTRAINTS> <CIRCLE> <THIN> </\discretionary{}{}{}CONSTRAINTS>}; at most 7 added special tokens, whose embeddings are warm-started from the mean pretrained embeddings of their natural-language names) --- and \textbf{text} --- a single comment line naming the active features in plain words (\texttt{\# design intent: circle thin}). Both are prepended to the front of the program text.

An evaluation entry is defined by its transpiled program: the prompt is a character prefix of that program and the scoring target is $\phi$ of that same program, so two entries whose transpiled code is byte-identical constitute one experimental unit however distinct their source histories. The clean pool is therefore deduplicated on the sha256 of the transpiled CadQuery program rather than on the source record, which reduces the 9,442 clean held-out programs to 4,641 unique programs, 3,834 of them carrying a non-empty four-tag intent. The choice of key is load-bearing rather than cosmetic: deduplicating on the model completion alone leaves 44 code-level duplicate clusters and 72 redundant entries in the pool, the largest cluster holding ten copies. The pool itself is reproducible --- the pool derived from the frozen cache of our initial evaluation and the pool recomputed from the raw dataset agree hash-for-hash on all 4,641 programs. Our \textbf{initial held-out set} predates this step and is reported throughout as such: its 100 entries cover 50 unique programs, 86 entries carry at least one active intent feature, and the 76 entries with a non-empty four-tag intent cover 38 unique programs spanning six of the eleven intent profiles.

A four-tag intent profile is the set of CIRCLE/NGON/THIN/TALL tags $\phi$ assigns to a program; the 3,834 non-empty-intent programs of the clean pool occupy eleven such profiles. The \textbf{replication sample} draws 400 of them under a stratification frozen before any evaluation ran: each profile is raised to a floor of fifteen programs, or exhausted if it holds fewer, and the remaining budget is allocated at a common sampling rate --- 9.49--9.71\% as realized --- with largest-remainder rounding and draw seed 20260805, so that the draw is bit-reproducible from the seed.

\begin{table}[ht]
\centering
\caption{\textbf{The eleven four-tag intent profiles of the clean held-out pool, and the stratified draw.} ``Initial'' counts the same profile's unique programs in the 76-entry / 38-program analysis subset of the initial held-out set.}
\label{tab:1}
\small
\adjustbox{max width=\columnwidth}{%
\small\setlength{\tabcolsep}{4pt}
\begin{tabular}{lrrrllrrr}
\toprule
\textbf{Profile} & \textbf{Pool} & \textbf{Drawn} & \textbf{Initial} & \textbf{} & \textbf{Profile} & \textbf{Pool} & \textbf{Drawn} & \textbf{Initial} \\
\midrule
THIN & 1,174 & 112 & 9 &  & CIRCLE+TALL & 112 & 15 & 0 \\
NGON+THIN & 736 & 70 & 4 &  & CIRCLE+\discretionary{}{}{}NGON+\discretionary{}{}{}THIN & 87 & 15 & 0 \\
CIRCLE+THIN & 609 & 58 & 9 &  & CIRCLE+NGON & 75 & 15 & 0 \\
CIRCLE & 412 & 40 & 3 &  & NGON+TALL & 70 & 15 & 0 \\
NGON & 411 & 39 & 11 &  & CIRCLE+\discretionary{}{}{}NGON+\discretionary{}{}{}TALL & 6 & 6 & 0 \\
TALL & 142 & 15 & 2 &  & \textbf{Total} & \textbf{3,834} & \textbf{400} & \textbf{38} \\
\bottomrule
\end{tabular}
}
\end{table}

Ten of the eleven profiles reach the fifteen-program floor; CIRCLE+NGON+TALL exhausts its pool at six, and the frozen protocol reports it while excluding it from every profile-level guardrail mean, so that a six-program profile cannot move a guardrail on equal footing with the 112-program THIN. All 400 programs carry a non-empty four-tag intent, so the analysis set is the full 400 with one scored entry per program and no within-cluster averaging step at all; the analysis script asserts that uniqueness rather than relying on it. The mean intent carries 1.54 of the four tags. Five of the eleven profiles --- 66 of the 400 programs --- have no representative whatsoever in the initial analysis subset, and only 2 of the 400 programs appear among that subset's 38, so the replication is substantially independent of the evaluation it replicates.

Stratification deliberately makes the sample's profile mixture differ from the natural mixture of the clean pool: TALL-bearing profiles are 51/400 = 12.8\% of the sample against 8.6\% of the pool. The main analysis leaves the 400 programs unweighted, so the population about which it draws inference is a \textbf{profile-balanced} one, and every aggregate claim in this paper is stated at that level. A reweighting of the same per-profile means to the pool's natural frequencies is computed and reported as a descriptive sensitivity --- it answers whether a conclusion is driven by the over-sampling of rare profiles --- and never enters a decision rule (§4.1).

The two held-out samples are reported side by side and are never pooled or placed on a common scale. Two differences make a cell-by-cell comparison of their absolute levels uninterpretable, and both were recorded before the replication sample was scored. The first is profile composition, which differs by construction, as Table 1 shows. The second is the generation-truncation regime: measured before launch under a matched condition --- same checkpoint, text header, correct content, 40\% prefix --- 21.0\% of generations on the new pool reach the completion cap against 4.5\% on the earlier pool. The cause is not longer ground truth (only 2.3\% of the new sample's ground-truth continuations exceed the cap) but a model that more often fails to emit EOS on this pool, most plausibly because the pool covers five intent profiles the earlier sample did not contain at all; under the execution-inclusive metric of §3.4 a truncated generation usually fails to execute and is scored zero. This is not a claim that one evaluation sits systematically above or below the other. The observed level differences run in mixed directions across arms --- the wrong-header arms are in fact higher on the replication sample --- which is precisely why the incommensurability is attributed to composition and truncation regime rather than to a uniform shift. Every number in this paper is accordingly labelled with the sample it comes from.

\subsection{Model M and the unconditioned baseline}

\label{sec:model_m_and_the_unconditioned_baseline}

\textbf{Model M} is the system under diagnosis: a conditioned generator built in the pattern of current LLM-fine-tuning practice (CADmium, FlexCAD, CAD-HLLM; §2.1--2.2) --- a pretrained code LLM fine-tuned on programs with a conditioning signal prepended, here produced automatically by $\phi$. Three design choices define it. First, M is trained exclusively on \textit{consistent} pairs $(h(\mathcal{I}_i), P_i)$: every header is derived from its own program and is thus consistent with it under $\phi$, so the training signal rewards learning the header$\leftrightarrow$program correlation --- M is given both the opportunity and the incentive to acquire the header$\rightarrow$program mapping whose causal role §4.2 tests. Second, the header participates in the loss (§3.1), so M models the header jointly with the program rather than treating it as fixed context. Third, M is instantiated under both header tokenizations (token and text, §3.2), so that conclusions about conditioning do not hinge on a particular surface form. The hypothesis M carries into evaluation is directly testable: if conditioning is content-sensitive, M's outputs should track header \textit{content} --- separating correct from wrong headers --- and not merely header presence.

Concretely, M fine-tunes Qwen2.5-Coder-1.5B \cite{hui2024qwen25coder}, a decoder-only Transformer code LLM, with QLoRA-style adapters \cite{dettmers2023qlora} --- the base model quantized to 4-bit (NF4, double quantization, bfloat16 compute), with LoRA on all attention and MLP projections (rank 16, $\alpha = 32$, dropout 0.05; $\approx$18.5M adapter parameters, \textasciitilde{}1.2\% of the base model --- the resized token embedding and LM head are additionally saved so the newly added header-token rows can train, with all pre-existing embedding rows gradient-frozen), batch size 4 with gradient accumulation 8 (effective 32), learning rate $2 {\times} 10^{-4}$ with linear decay over 3 epochs, gradient clipping at norm 1.0, 3 seeds per condition. The \textbf{unconditioned baseline} is trained identically but with no header --- the never-header-trained reference for every wrong-vs-none contrast. A \textbf{CADCON-CFG} variant, trained with classifier-free-guidance-style header dropout \cite{ho2022cfg} so that unconditional fallback is representable at inference, was never re-evaluated under the geometric metric and supports no claim made here; its configuration and motivation are given in Appendix D. Table 2 lists the full configuration; it covers M, the baseline, and the derangement control R (§3.5), which differs from M \textit{only} in the header$\leftrightarrow$program pairing of its training data.

\begin{table}[ht]
\centering
\caption{\textbf{Hyperparameters.} All rows except the last are shared by M, the unconditioned baseline, and control R (§3.5) --- R and M are trained \textit{without} header dropout, and R differs from M only in the header$\leftrightarrow$program pairing of its training data; the last row's GRPO cells configure the study reported in Appendix D and its CFG cell applies to the CADCON-CFG variant only.}
\label{tab:2}
\small
\begin{tabularx}{\columnwidth}{XXXX}
\toprule
\textbf{Hyperparameter} & \textbf{Value} & \textbf{Hyperparameter} & \textbf{Value} \\
\midrule
Base model & Qwen2.5-Coder-1.5B & SFT epochs & 3 \\
LoRA rank / $\alpha$ / dropout & 16 / 32 / 0.05 & Max sequence length & 1,024 \\
SFT learning rate & $2{\times}10^{-4}$ (linear decay) & Completion cap (calibrated) & 474 at 40\% prefix (750 at 0\%) \\
Batch $\times$ grad-accum & 4 $\times$ 8 (eff. 32) & Eval prefix fraction & 0.40 (character-level) \\
Seeds (SFT conditions) & 3 & Training programs & 10,000 \\
Gradient clip norm & 1.0 & Held-out samples & 400 (replication) / 100 (initial) \\
Eval batch size (frozen) & 8 & Hardware & 2$\times$ RTX 3090 (24 GB) \\
GRPO lr / steps / gens & $5{\times}10^{-7}$ / 100 / 4 & CFG header-dropout prob. (variant only) & 0.50 \\
\bottomrule
\end{tabularx}
\end{table}

\subsection{Execution-aware validation on the produced geometry}

\label{sec:execution_aware_validation_on_the_produc}

The same deterministic extractor $\phi$ defines both the training-time headers and the adherence metric our initial evaluation used, applied to the ground-truth program to obtain the target and to the generated program \textit{text} to obtain what was produced. A metric of that shape asks whether the generation carries the surface pattern $\phi$ searches for --- the very pattern from which the header was built and which the model was trained to emit beneath that header. It cannot separate a model that realized the requested geometry from one that reproduced the token pattern the header names, and it credits both equally. A single-condition audit in the initial evaluation could bound the resulting inflation but not remove it, because bounding it requires assuming what fraction of matched patterns correspond to realized geometry, which is the quantity in question. This section describes the instrument that removes the assumption; what it changes about the reported benefit of conditioning is a result and is given in §4.4.

Adherence is therefore scored on the produced \textit{solid} rather than the produced text. Each generated program is executed in an isolated CadQuery subprocess under a 10 s timeout; on the resulting B-rep solid, five executable assertions read the realized features --- CIRCLE from a circular edge, NGON from a face with $\geq$5 straight edges, THIN and TALL from the executed bounding-box aspect ratio, MULTI\_PART from the solid count --- grouped per requirement in the style of CADTests \cite{mallis2026cadtests}. The assertion module shares no code with $\phi$: it never inspects program text, and $\phi$ never inspects a solid. That independence is of implementation and of the object scored, not of every parameter --- THIN and TALL reuse $\phi$'s 0.3 and 2.0 aspect-ratio cut-offs, applied now to the executed bounding box rather than to the program text --- and the pre-registration records the consequence: in the audit of the initial evaluation the two detectors agreed sample-for-sample on all 62 executable generations, so what the geometric metric adds is the execution gate and the scoring of a solid rather than a string, not disagreement between two detectors on the same executable output. The intent target remains $\phi(\text{GT})$, which is what defines the task; only the produced side changes, which is exactly where the circularity lived. What this instrument removes is therefore produced-side circularity specifically, and the section is named for that rather than for the absence of circularity in general.

Executing the 400 ground-truth programs themselves puts a scale on both the agreement between the two detectors and the metric as a whole, and we report it because without it no absolute arm level in this paper is interpretable. The two detectors agree on 80.2\% (CIRCLE), 79.2\% (NGON), 68.8\% (THIN) and 94.0\% (TALL) of the 400 programs, rising to 97.0 / 84.5 / 87.5 / 94.8\% on the 271 that execute; almost all of the residual is $\phi$ reading a feature off program text that the executed solid does not realize, and $\psi$ reads a feature $\phi$ missed on 17 programs for NGON and 13 for TALL. The per-tag confusion matrix is in Appendix B. More consequentially, ground truth scores only 0.567 on its own four-tag target under this metric: 156 of the 400 reference programs realize none of the four features $\phi$ ascribes to them, 129 of those because they do not execute at all, and only 210 realize their intent in full. This is not a ceiling on what a model can score, and we say so explicitly because the temptation to read it as one is strong. The arms complete their own programs from a 40\% prefix and frequently run where the transpiled reference does not: on the 129 programs whose reference fails to execute the arms reach 0.34 to 0.47, and the correct-header text arm scores a full 1.0 on 38.9\% of the sample--seed cells where the reference scores zero. What the figure does fix is the meaning of the target. A program is scored against $\phi$'s reading of the reference \textit{text}, not against a property the reference \textit{solid} realizes, so an absolute adherence level in §4 measures agreement with a regex-declared attribute set rather than the fraction of a design intent achieved, and the task is adherence to synthetic geometric attributes rather than to human-annotated design intent.

The composition of the reference set bears on effect sizes as well as on levels. Every arm is scored against the same target, so no paired comparison changes sign, but the wrong-versus-baseline contrast is about twice as large where the reference realizes its full intent as where it realizes none: text $-0.122 / -0.103 / -0.158$ on the 210 against $-0.016 / +0.001 / -0.060$ on the 156, and token $-0.211 / -0.201 / -0.217$ against $-0.091 / -0.049 / -0.119$. The aggregate of §4.1 is a mixture over those strata, which is the same arithmetic that makes the exclusion analyses of §6 mostly compositional.

A program that fails to execute realizes no features and scores zero on every requirement --- execution-inclusive scoring. Executability here is a property of the scoring subprocess and not of the in-training-loop CadQuery oracle, whose validity bit is reported separately and enters no decision rule. Execution-inclusive scoring is the conservative convention for the claims made here, since a generation that does not run cannot be credited with an intent it never realized; but it couples the primary metric to executability, and a wrong-header effect could in principle be an executability effect reported as an adherence effect. Every decision-critical contrast is therefore reported a second time on the subset where the compared arms both execute, with the zero-filling removed entirely (§4.1, §4.2); for the control's dissociation test that executable-intersection re-analysis is a pre-registered guardrail rather than a check invented after the fact (§3.5). That subset conditions on executability, which is itself an outcome the header arm affects, so it is descriptive rather than decisive; the evidence that carries the point instead is how often each arm produces a runnable program at all, reported with the arm levels in §4.1. The same caveat attaches to every executable-conditioned quantity in this paper, wherever it points --- the four-arm intersection behind the control's guardrail (§4.2), the control's executable-only arm levels, the token header's benefit where both arms run (§4.4), and the executable-only rows of the wrong-target table (§4.2). We apply the qualification uniformly rather than only where it costs us.

MULTI\_PART is reported but excluded from the primary metric, which is a four-tag mean per-sample positive-label recall over CIRCLE, NGON, THIN and TALL --- the mean over samples of the fraction of a program's intended features that are realized, written ``adherence'' below. The name matters because the quantity is not a macro-average of per-class recalls and, being a recall over the intended features only, carries no penalty for a feature the model realizes that the intent did not ask for; §4.4 reports precision, $F_1$ and exact-set accuracy beside it for that reason. Writing $T$ for those four tags and $\psi(\hat{P})$ for the features the executable assertions read off the solid produced by a generated program $\hat{P}$, adherence on one sample is

\[
\mathrm{adh}(\hat{P}) \;=\; \frac{\sum_{t \in T} \mathcal{I}_t \, \psi_t(\hat{P})}{\sum_{t \in T} \mathcal{I}_t} ,
\qquad
\psi(\hat{P}) = \mathbf{0} \;\; \text{if } \hat{P} \text{ does not execute} ,
\]

with the intent $\mathcal{I} = \phi(P)$ read from the ground-truth program. A sample whose four-tag intent is empty carries no requirement and is excluded; every adherence figure reported under the geometric metric is the mean of $\mathrm{adh}$ over the samples of a cell. The training-side regex recall reported beside it (§4.4) is the same expression with the produced side read by $\phi$ off the generated text rather than by $\psi$ off the solid, and without the execution clause. The second clause is execution-inclusive scoring stated formally: a program that does not run contributes a zero, not a missing value. The exclusion of MULTI\_PART is a property of the geometry rather than of the model: overlapping extrudes fuse into a single solid on execution, so an intent that the extractor reads off the program text as multi-part is frequently not realizable as a solid count at all. In the pre-registration audit of the initial evaluation, 6 of 26 executable ground-truth MULTI\_PART entries realize more than one solid and 0 of 62 generated ones do; on the replication sample, 7 of 3,626 executable generated samples carrying a MULTI\_PART intent realize more than one solid. A requirement that ground truth itself satisfies less than a quarter of the time measures the transpiled representation, not the generator.

Because the primary metric is a recall, a model could in principle inflate it by emitting every feature. Measured on the replication sample it does not, and could not do so unobserved: across the nine evaluated arms, executable outputs realize on average 0.50--1.23 distinct primary features per sample (unconditioned baseline 1.23; header-informed arms 0.99--1.23; masked arms $\approx$0.50), against a mean intent size of 1.54 tags, and conditional on realizing at least one feature the average is 1.23--1.31 in every arm. No arm approaches the four features that feature-spamming would require, and the arm ordering the paper reports is not reproduced by the number of features emitted.

Both metrics can be computed on the same generations, which makes their difference a measurement of the instrument rather than of the model. The training-side regex recall is therefore reported beside the geometric recall throughout, under a convention-matched definition of four tags over the same programs. The two then differ in two respects rather than one: where they look, and whether the generated program has to run at all --- the regex scores a string whether or not it executes, while the geometric metric scores a solid and gives a program that does not run a zero. Separating those two respects is what §4.4 does, because the separation turns out to carry the result. That comparison was specified in advance, and its outcome is reported in §4.4 because it is the quantity a reader needs in order to weigh conditioning claims made elsewhere in this literature under signal-derived metrics. The regex figure never enters a decision rule: every pre-registered decision in this paper rests on the geometric metric alone.

\subsection{The derangement control R}

\label{sec:the_derangement_control_r}

A wrong-header harm (§4.1) is causally ambiguous: it could show that the model reads the header's content, or merely that a mismatched header is an out-of-distribution input for a header-trained model. The control that separates these must experience wrong headers as \textit{in-distribution} while having had no consistent header$\leftrightarrow$program relationship to learn from. We construct it by \textbf{derangement training}. \textbf{Model R} is a derangement-trained negative control, designed to satisfy three properties at once: (i) \textit{identical training recipe} --- the same base model, LoRA configuration, training programs, step count, and seeds as M (Table 2), so a behavioral difference between the two cannot be attributed to the training recipe --- architecturally, R is indistinguishable from M; (ii) \textit{identical header marginal} --- every training program receives \textit{another} program's header under a fixed derangement plan, so every header string R sees is byte-identical in form and frequency to what M sees, and a wrong header at evaluation is not a rare or novel input for R; (iii) \textit{no learnable mapping} --- the derangement destroys the header$\leftrightarrow$program correlation, so R cannot systematically acquire the mapping whose causal role is under test. The destruction is not perfect: the small feature vocabulary leaves 8.8--9.7\% of deranged training headers correct-by-chance (measured per seed on the 10,000 training programs: 8.8\% / 9.7\% / 9.2\%), a residual that biases R toward content-\textit{sensitivity} --- making any observed immunity of R conservative. If the model's response to a wrong header is a marginal/mechanical perturbation effect, R --- a model for which uncorrelated headers are the training distribution itself --- should be moved by wrong headers too; if that response requires the learned header$\rightarrow$program mapping, R should be unmoved. What the contrast can decide is therefore the content sensitivity, not the level at which either model's wrong arm sits relative to the unconditioned baseline (§4.2). Both predictions are evidential only if R remains competent, which the pre-registered gate below tests; the gate's outcome, and the sense in which it carries only on its $\geq$2/3-seed rule, are reported in §4.2 and their consequences for the reading in §6.

The control protocol pre-registers, in order: (i) a \textbf{competence gate} --- R must stay non-trivially above the 40\%-prefix floor, with per-seed execution-inclusive four-tag adherence $\geq$ 0.30 in \textit{both} header arms on at least two of three seeds, and executability within 0.10 of the unconditioned baseline's --- which guards against a spurious dissociation manufactured by a broken control, and whose 0.30 floor was chosen ex ante with the initial evaluation's unconditioned baseline of $\approx$ 0.42 known and deliberately undercut, because a correlation-deprived control is expected to be weaker than an unconditioned model; (ii) the \textbf{dissociation test} (Test D) --- the per-sample interaction between M's and R's correct$\rightarrow$wrong adherence drop, paired Wilcoxon per seed, requiring $\geq$2/3 seeds positive and significant with a reverse-direction veto; and (iii) three guardrails --- an executable-intersection re-analysis that removes the execution-inclusive zero-filling entirely, a profile-level direction check, and a non-zero-pair count that flags any seed with too few non-zero paired differences (thresholds in §3.6). The protocol additionally pre-registers a supporting-only, content-graded dose-response probe on M --- a \textbf{corrupted} arm (one or two primary features falsified) and a \textbf{random} arm (well-formed, size-matched, sample-unrelated feature sets; designated non-evidential ex ante) --- whose criterion and descriptive outcome are reported in §4.2. The gate's thresholds carry the interpretation of the outcome and not merely its validity: the frozen outcome mapping sends a failed gate to ``control invalid'', requires the conservative headline in that case, and forbids the phrase ``dissociation confirmed'' --- a mapping fixed before either of R's evaluations existed. A built-in assertion re-reads the frozen scores of the corresponding wrong-vs-none evaluation and checks that M's arm means match the published values --- a wiring/anti-drift \textbf{integrity check} (it re-reads the same frozen artifacts, so it is not an independent reproduction).

R was evaluated twice: first on the initial held-out sample, and then on the 400-program profile-stratified sample that carries the primary result, under a frozen increment that changed which programs are scored and nothing else. The control model itself was not retrained for the second evaluation --- the same three derangement-trained checkpoints were re-evaluated, and M's arms were reused from the primary evaluation, so the re-evaluation consists of six additional pure-inference jobs. Both runs are reported (§4.2), with the 400-program run primary. The increment's single permitted change, its two gating assertions, and the blinding conditions under which it was frozen are described in §3.6.

\subsection{Pre-registration, freezing, and the blinded power check}

\label{sec:pre_registration_freezing_and_the_blinde}

The initial evaluation was a \{0\%, 40\%\} program-prefix $\times$ \{correct, wrong, masked\} header matrix, for both tokenizations, over three seeds, plus the never-header-trained unconditioned baseline at both prefixes --- (6 header cells $\times$ 2 tokenizations + 2 baseline cells) $\times$ 3 seeds = 42 pure-inference jobs on the frozen 100-entry held-out set. The replication re-runs the 40\%-prefix column on the profile-stratified 400-program sample: (3 header cells $\times$ 2 tokenizations + 1 baseline cell) $\times$ 3 seeds = 21 pure-inference jobs, all 400 programs scored in every cell. The control adds 18 jobs in its first run and 6 in the re-evaluation (§3.5). Generation is batched greedy decoding with no sampling, up to the calibrated completion cap of Table 2 --- 474 tokens at 40\% prefix, derived as the 95th percentile of the completion length of twenty training programs, a small calibration sample whose consequences are recorded in §6. The \textbf{wrong} header is a well-formed header taken from another sample under a derangement plan constructed once per evaluation from a fixed rng seed, so the arm receives real headers carrying incorrect content rather than malformed ones; the identical plan is applied to M and to R, and the re-evaluation asserts that the injected header is byte-identical between the two models at every (sample, seed) before any statistic is computed. The \textbf{masked} header suppresses the header at inference for a model trained with one.

The evaluation batch size is an experimental parameter here, not a throughput setting, and is frozen accordingly. Greedy decoding is not batch-invariant in this implementation: the same prompts decoded at batch 8 and batch 32 agree byte-for-byte on only 78\% of generations at 40\% prefix. A control experiment locates the cause in left-padding rather than in the batch dimension --- a set of prompts of exactly equal length, which produce no padding at all, decode identically across the two batch sizes on 32 of 32 generations, under a long generation cap that simultaneously rules out the ``longer generations diverge more readily'' explanation. That control changes the model, the prefix regime and the prompt diversity at the same time as it removes the padding, so it identifies a mechanism rather than bounding a magnitude, and we do not claim more from it. Fixing the batch size and the sample order fixes the padding and makes decoding reproducible; it does not by itself exclude an interaction between padding and treatment, since header length differs by arm and therefore so does the left-pad width. What bounds that residual is a measurement on the primary metric rather than on the text, reported in Appendix B: changing the batch size moves the per-sample score on 2 of 64 samples and the arm mean by at most 0.0234, in a direction that reverses between batch sizes. The batch size is 8 for every arm and every job of both evaluations, and the frozen protocol forbids changing it once a run has started.

Decision rules were frozen with SHA-256 hashes before any scoring in every run. The unit of analysis is one unique transpiled program; the replication sample makes this automatic by construction and the analysis script asserts it rather than relying on it. Every difference is same-seed paired --- $\Delta(s)$ is the conditioned arm at seed $s$ minus the baseline at seed $s$, matched per program --- and seed-averaged baselines are forbidden. The frozen test specification is \texttt{scipy.stats.wilcoxon(d, zero\_method=``pratt'', alternative=\ldots{}, method=``asymptotic'')} at $p<0.05$, with $p$ reported unrounded, an all-zero difference vector counted as non-significant, and direction taken from the sign of $\mathrm{mean}(d)$. The criterion for a wrong-vs-none cell requires all of: $\geq$2/3 seeds in the predicted direction \textit{and} $\geq$2/3 seeds significant on the text side; $\geq$2/3 seeds in the direction on the token side; a reverse-direction veto, under which a significant result in the opposite direction on any seed in either tokenization fails the criterion outright; and a profile-level direction guardrail on the mean of the per-profile effects (intent profiles: the distinct four-feature intent combinations present in the sample --- six in the initial held-out set, eleven in the replication sample, where the guardrail averages the ten with $n \geq 15$ and the eleventh, exhausted by its pool at six programs, is reported only; Appendix C). A non-zero-pair guardrail flags any seed whose count of non-zero paired differences falls below 60 ($=0.15n$ on the replication sample; 15 on the initial one), and a strict reading fixed in advance fails the criterion if its $\geq$2/3 majority does not survive removing the flagged seeds. Because the profile guardrail acts on a \textit{mean}, it licenses nothing about any individual profile: the protocol pre-registers no per-profile test, p-value or power, and per-profile estimates are descriptive throughout.

The 40\% character-level prefix sits below the median first occurrence of the five feature signatures (52--87\% of program length), but the tail of that distribution is not negligible and its weight depends on which intent profiles the sample contains. On the profile-stratified 400-program sample, 17.8\% of programs already satisfy at least one of their four-tag intent features inside the prefix, and 18.0\% counting MULTI\_PART; on the earlier, unstratified sample the corresponding figure was 8.0\%. The increase is a foreseeable consequence of the stratified draw, which introduces the CIRCLE-family profiles the earlier sample largely lacked, and \texttt{.circle(} occurs early in a program. Leakage is symmetric across conditions --- every arm at a given index receives the identical prefix --- so it cancels in the paired differences that carry every decision here; its effect is to raise the recall floor and convert samples into exact ties, which costs power rather than introducing bias (§6). The prefix fraction was a frozen parameter and was not adjusted in response to this measurement, which was recorded before any decision criterion was computed.

Four protocols were frozen in sequence. The first three were frozen before any of the data they govern existed; the fourth, the Test D increment, was frozen before the control's data on this sample existed but after M's arms had been scored, an asymmetry set out at the end of this section. The six-cell protocol --- decision rules, tests, thresholds --- was frozen with SHA-256 hashes before any scoring. The control protocol and its analysis script (\texttt{f4\_analysis.py}, sha256 \texttt{c506ab4e\ldots{}}) were frozen on 2026-07-07, before the control model's evaluations ran on 2026-07-08. No rule, threshold or test was altered after freezing in either.

The replication protocol --- decision rules, thresholds, the frozen 400-program list with its eleven profile definitions and per-profile quotas, the job manifest, and the analysis script --- was frozen with SHA-256 hashes (protocol \texttt{b630efae\ldots{}}) at 02:25 UTC on 2026-08-06, and the evaluation launched at 02:37 UTC the same day; the pool construction, the stratified draw and its seed all precede the freeze. That protocol also fixes, before unblinding, the single circumstance under which the sample could have been enlarged, because the difference between a legitimate sample-size rule and optional stopping lies entirely in timing. After geometric scoring and before the decision analysis ran, a separate blinded script counted the non-zero paired differences per seed in the decision cell. The script computes counts only, and was verified by AST inspection to contain no mean, no sign, no direction and no Wilcoxon path, so the blinding is enforced by the tool rather than by the analyst's discipline. Had at least two of the three seeds fallen below 60 non-zero pairs in either tokenization, the run would have been reported as underpowered and inconclusive, and any enlargement would have constituted a separate, independently pre-registered experiment at $n = 800$, never pooled with this one; had it not, the analysis would be final at $n = 400$ and enlargement forbidden thereafter, whatever the results turned out to be. The check ran before any effect direction, mean or p-value existed and returned no trigger: 123 / 113 / 126 non-zero pairs per seed on the text side and 123 / 116 / 126 on the token side, roughly twice the threshold in every cell. The analysis proceeded at $n = 400$, and no expansion was permitted afterwards. The same protocol fixes in advance, in full sentences, how the paper would be written under each outcome --- including the failure case, which it requires to be reported as a powered clean negative rather than softened into an inability to confirm --- and forbids the two write-ups from being exchanged after the fact or a third being introduced.

The Test D increment was frozen before any of the control's data on the replication sample existed. It changes the analysis sample and nothing else: the decision rules, the competence-gate thresholds (0.30 adherence floor, 0.10 executability margin), the test specification, the reverse-direction veto and the outcome mapping are carried over verbatim from the frozen control protocol. Three quantities scale with the sample --- the analysis set (400 unique deduplicated programs, requiring no within-cluster averaging), the profile set (eleven four-tag profiles, with the direction guardrail counting the ten that have $n \geq 15$), and the non-zero-pair guardrail (60 $=0.15n$) --- and the increment's supplementary clause requires both control runs to be reported together, with the replication run primary, forbidding the reporting of only the favourable one.

Blinding at that freeze was asymmetric, and we state which side was open. M's correct- and wrong-header arms on the replication sample had already been scored as part of the primary result, and their per-seed and per-profile values had been inspected and recorded in the run log; R's arms on that sample did not exist. Because the test statistic is $\mathrm{drop}_M - \mathrm{drop}_R$ and $\mathrm{drop}_R$ was unobserved, the outcome of the test was not knowable at freezing time, so the re-evaluation is prospective in the sense that matters for Test D --- but it is not prospective with respect to M, and we do not present it as such. The increment's sha256 (\texttt{0a5ff59d\ldots{}}) and its commit timestamp allow the ordering to be verified independently. Two assertions gate the analysis before any statistic is computed: M's arm means must reproduce the frozen replication verdict, and M and R must have received the byte-identical injected wrong header at every (sample, seed), without which the interaction would be meaningless.

All four pre-registration documents, their hash manifests, the frozen held-out lists and profile definitions, and every job manifest are part of the release (§ Reproducibility and release); the full freeze chain, including one procedural event during the re-evaluation, is recorded in Appendix A.

\section{Results}

\label{sec:results}

We report the geometry-side metric (§3.4) as primary; the training-side regex appears beside it in §4.4 to calibrate the instrument rather than to decide anything. All adherence numbers are four-feature execution-inclusive mean per-sample positive-label recall on the 400-program profile-stratified held-out sample (§3.2), averaged over three seeds unless a per-seed breakdown is given. Every program in that sample carries a non-empty four-tag intent; the eleven profiles range from $n = 6$ to $n = 112$, and the pre-registered profile guardrail counts the ten with $n \geq 15$. The inferential population is the profile-balanced one the stratified draw defines; the reweighting to the profiles' natural frequencies in the clean held-out pool is a report-only sensitivity and decides nothing. Where our earlier, smaller evaluation is cited it is named as such: the two runs differ in intent-profile composition, by construction, and in the generation-truncation regime (§6), so their absolute levels are not commensurable and are never pooled or placed on a common scale. Per-profile effects appear throughout and are descriptive without exception --- the frozen analysis defines no per-profile test, p-value or power, and the profile-level guardrail is a direction check on the mean of the per-profile effects, which licenses nothing about any individual profile (§4.3; Appendix C). Models: \textbf{M} (standard, informative-header-trained) and \textbf{R} (derangement control, §3.5). Header arms at evaluation: \textbf{correct}, \textbf{wrong} (another program's ground-truth header, injected at evaluation), \textbf{masked} (header suppressed at inference), and the \textbf{unconditioned baseline} (never header-trained).

\subsection{A wrong header is worse than never conditioning}

\label{sec:a_wrong_header_is_worse_than_never_condi}

At 40\% prefix the ordering under the independent metric is correct > never-conditioned > wrong on the text header; on the token header the correct arm does not clear the baseline at the program level (§4.4), while the wrong-below-baseline leg holds under both tokenizations. The never-header-trained baseline reaches 0.406, against 0.325 for the text wrong-header arm and 0.254 for the token one.

Every decision-critical contrast is reported at two units of analysis, because the 400 programs present only 298 distinct 40\% prefixes and programs sharing one are not independent draws (§6). At the program level the per-seed paired effect is negative on all three seeds under both tokenizations and significant on all six of the resulting cells: text $-0.072 / -0.056 / -0.115$ (same-seed paired Wilcoxon, \texttt{pratt}, one-sided; $p = 2.5{\times}10^{-3} / 0.011 / 9.1{\times}10^{-7}$), token $-0.155 / -0.132 / -0.170$ ($p = 1.3{\times}10^{-11} / 5.3{\times}10^{-9} / 2.2{\times}10^{-13}$). At the level of distinct model inputs --- the frozen test specification re-run unchanged after averaging each prefix cluster's paired differences to one value --- text is $-0.072 / -0.041 / -0.061$ ($p = 9.9{\times}10^{-3} / 0.098 / 8.4{\times}10^{-3}$) and token is $-0.130 / -0.091 / -0.122$ ($p = 4.3{\times}10^{-8} / 1.6{\times}10^{-4} / 4.6{\times}10^{-7}$). Five of the six cells survive; text on seed 1 moves from $p = 0.011$ to $p = 0.098$ and would not be called significant. The pre-registered rule asks for two of three seeds per tokenization, which both units meet, so the frozen outcome mapping returns REPLICATED on either; what the cluster unit changes is the strength of the statement that can be made about the text header on a single seed, and it is the unit at which this paper states the result. No cell triggers the pre-registered reverse-direction veto.

The clustering is not uniform, and the largest collision class carries a disproportionate share of the program-level figure: 41 programs share a single input, and that one class supplies 31\%, 27\% and 48\% of the text-side effect across the three seeds, and 27\%, 30\% and 32\% of the token-side effect. Removing it leaves the program-level effect at $-0.055 / -0.046 / -0.067$ ($p = 0.017 / 0.036 / 1.4{\times}10^{-3}$), still negative and significant on 3/3 seeds; at the cluster unit the same 41 programs are, by construction, one observation out of 298. This is the concrete reason the cluster unit rather than the program unit carries the claim.

A hierarchical bootstrap over prefix clusters and seeds puts intervals on the same quantities. Pooled over the three seeds the wrong-versus-baseline effect is $-0.058$ (95\% CI $[-0.089, -0.026]$) for the text header and $-0.115$ ($[-0.147, -0.081]$) for the token one, both excluding zero; the Test D interaction of §4.2 is $+0.180$ ($[+0.146, +0.213]$). Per seed the text-side cluster intervals are $[-0.120, -0.025]$, $[-0.089, +0.007]$ and $[-0.109, -0.014]$, so the seed that loses significance under clustering is exactly the one whose interval covers zero. With three training seeds the seed level of that bootstrap rests on three observations and the pooled intervals should be read accordingly. These intervals were not pre-registered and decide nothing; they are reported because a paper that pre-registers every decision rule should not leave its effect sizes without uncertainty.

The pre-registered profile-level direction guardrail passes at a mean profile effect of $-0.046$, equally weighted over the ten profiles with $n \geq 15$. Two report-only reweightings of the same per-profile means bound the sensitivity of that figure to the draw: equal weight over all eleven profiles gives $-0.054$, and weighting by the profiles' natural frequencies in the clean held-out pool gives $-0.086$. The aggregate harm is therefore not an artifact of which profiles the stratified draw over-sampled --- under natural frequencies it is larger, not smaller. That is the only generality about the profile dimension this design licenses; §4.3 reports the heterogeneity it does not license claims about.

The masked arm also falls below the baseline, at 0.243 (text) and 0.238 (token), with direction and significance on 3/3 seeds under both tokenizations at the program level (text $-0.130 / -0.181 / -0.178$, $p \le 1.4{\times}10^{-8}$; token $-0.138 / -0.171 / -0.195$, $p \le 8.2{\times}10^{-10}$; profile guardrail $-0.069$); at the level of distinct model inputs the bootstrap intervals exclude zero on all six cells (text pooled $-0.073$, $[-0.107, -0.039]$; token $-0.079$, $[-0.110, -0.047]$). This cell was pre-registered as a second anchor and passed. A model that has been trained with a header and is then denied one is worse, under an evaluation the header played no part in defining, than a model that never had one --- the signature of learned header-dependence, here confirmed on the produced geometry rather than on the extractor that built the headers.

Within a single model the ordering runs the other way, and we state it rather than leave the title to imply otherwise. The wrong-header arm sits \textit{above} the masked arm, 0.325 against 0.243 on the text side and 0.254 against 0.238 on the token side. ``Worse than never conditioning'' therefore means worse than a separately trained never-header model throughout this paper, and never means worse than the same model with its header suppressed. Masking is not a clean null mode in any case: it hands a header-trained model an input it never saw in training, and the arms it produces are the most executable in the study --- 0.855 (text) and 0.828 (token) against the baseline's 0.603 --- while 59\% of those executable outputs realize no intent feature at all. Suppressing the header buys executability at the cost of content: the majority of what runs realizes none of the intent. A model with a represented unconditional mode, trained in rather than improvised at inference, is the comparison this design lacks (§5).

Execution-inclusive scoring gives a program that fails to run a zero on every requirement, so a wrong-header effect could in principle be an executability effect reported as an adherence effect. The evidence that bears on this without conditioning on an outcome is the executability of the arms themselves: the text wrong-header arm runs \textit{more} often than the never-header baseline, 0.638 against 0.603 pooled and higher on two of three seeds, so on that side the deficit cannot be a matter of programs that fail to run. The token wrong arm is at 0.507, below the baseline, and there an executability contribution is not excluded. A second check restricts the comparison to the subset on which the baseline and the wrong-header arm both produce executable geometry; the deficit persists on every seed under both tokenizations and is significant on all six: text $-0.118 / -0.138 / -0.118$ on $n = 183 / 194 / 200$ paired programs ($p = 4.0{\times}10^{-5} / 1.9{\times}10^{-6} / 1.7{\times}10^{-5}$), token $-0.152 / -0.136 / -0.147$ on $n = 154 / 157 / 179$ ($p = 2.2{\times}10^{-6} / 1.1{\times}10^{-5} / 1.5{\times}10^{-6}$). That subset is selected on a post-treatment variable --- executability is itself an outcome the header arm affects --- so it is a descriptive statement about the samples that happened to run in both arms and not a demonstration that execution plays no part. We report it at that strength.

Under greedy decoding many held-out programs are byte-identical across arms and tie, and three further mechanisms --- extractor truncation, prefix leakage and prefix duplication (§6) --- add ties symmetrically across arms. We therefore report the non-zero paired differences that carry each test rather than the nominal $n$. For the wrong-vs-baseline contrast these number 123 / 113 / 126 of 400 on the text side and 123 / 116 / 126 on the token side, against a pre-registered non-zero-pair guardrail of 60, and no seed was flagged underpowered. The harmful direction dominates in every cell (harmful / helpful / tied: text 76/47/277, 68/45/287, 89/37/274; token 98/25/277, 88/28/284, 103/23/274). These counts were produced by a blinded script --- verified by AST inspection to contain no mean, direction or Wilcoxon code path --- and executed before any effect direction or p-value existed; the check returned no trigger, and the protocol permits no enlargement of the sample thereafter (§3.6).

Our earlier evaluation reached the same ordering on a smaller and unstratified held-out set. Recomputed on the 38 unique programs its 76 scored entries cover --- averaging each program's paired differences to one value, as §6 does for the replication --- the wrong-vs-none effect there is $-0.079 / -0.132 / -0.151$ on the text side ($p = 0.118 / 0.064 / 0.023$, on 12 / 13 / 13 non-zero clusters): direction on 3/3 seeds and significance on 1/3, against a pre-registered probability of 0.463 that at least two of three seeds would reach significance at that unit. The earlier run therefore settled the question in neither direction, which is what the pre-registered replication was written to fix. The present run is the primary one, and the two runs' arm levels are kept apart throughout (§6).

What the result establishes is narrow in one respect and consequential in another. Narrowly, it is a statement about one checkpoint, one program representation and one conditioning interface. Consequentially, it removes the assumption a designer of such an interface would otherwise make: that a conditioning field is at worst inert, so that a stale, mistaken or upstream-inferred intent specification costs at most the benefit conditioning would have brought, leaving the model where an unconditioned model would have been. Under an evaluation that shares no code with the extractor defining the header, on a sample drawn to span every intent profile the corpus presents, and in the regime in which such a model would actually be used --- extending a partially written program --- a well-formed header carrying another program's intent leaves the model measurably further from the true design intent than never training the field at all. In this setting a wrong conditioning signal is not inert, and the masked arm shows that the model's reliance on the header field is itself acquired during conditioning rather than present beforehand. What the ordering does not yet say is \textit{why} the wrong header hurts, which is the question §4.2 tests.

\subsection{The content sensitivity requires the learned header$\rightarrow$program mapping}

\label{sec:the_content_sensitivity_requires_the_lea}

The result of §4.1 admits two readings. Either the model reads the header's content, so that incorrect content misdirects generation, or a mismatched header is simply an off-distribution input for a header-trained model and would perturb it whatever it said. The derangement control R separates them (§3.5). R saw a byte-identical header marginal during training but no consistent header$\leftrightarrow$program relation, so an uncorrelated header is R's training distribution rather than an anomaly, and R had nothing from which to acquire a header$\rightarrow$program mapping. If the harm is mechanical, R should suffer it too; if it requires the learned mapping, R should be immune, provided R is competent enough for its immunity to mean anything.

The control was first evaluated on the earlier held-out sample and then re-evaluated, under an increment protocol frozen before any of its data on the present sample existed, on the same 400 deduplicated profile-stratified programs that carry §4.1. The increment changed the analysis sample and nothing else: decision rules, gate thresholds, test specification, reverse-direction veto and outcome mapping were carried over verbatim, and only the analysis set, the profile set and the non-zero-pair guardrail scale with $n$ (§3.6). Both runs are reported, with the 400-program run primary. Except where the earlier run is named, every number below is from the replication sample: text headers, 40\% prefix, three seeds.

R clears the pre-registered competence gate, and the margin by which it does so is part of the result rather than a footnote to it. Its execution-inclusive four-feature adherence is 0.267 / 0.354 / 0.407 per seed under a correct header and 0.273 / 0.338 / 0.410 under a wrong one. The gate requires at least 0.30 in both arms on at least two of three seeds: seeds 1 and 2 clear the floor in both arms, seed 0 falls below it in both, and the gate therefore carries on the $\geq 2/3$ rule rather than unanimously. R's executability is 0.547 pooled over its two arms against the baseline's 0.603, inside the pre-set 0.10 margin (threshold 0.503); per seed it is 0.417 / 0.605 / 0.619 against the baseline's 0.583 / 0.578 / 0.650, so the pooled figure on which the frozen gate is defined absorbs a 0.17 shortfall on seed 0. The protocol was written expecting R to land near the unconditioned baseline, at roughly 0.42; the observed control sits well below that expectation. On its own executable outputs R's adherence is 0.628 (correct arm, $n = 655$) and 0.621 (wrong arm, $n = 658$) against the baseline's 0.673 ($n = 724$) --- below baseline rather than indistinguishable from it (Mann--Whitney two-sided, pooled over R's arms, $p = 0.015$; correct arm alone $p = 0.054$) --- and far below M's correct-header 0.836 ($n = 751$; $p = 1.2{\times}10^{-22}$ against R's correct arm). On the earlier sample the corresponding comparison was indistinguishable from baseline; on this one it is not. R is accordingly an impaired control rather than a floored one, and the dissociation below must be read as a contrast between a model that has a header$\rightarrow$output causal path and one that does not, not between two models of matched ability.

That impairment has a specific structure, and it bounds what the control can license. R sits below the unconditioned baseline under a \textit{correct} header by $-0.063$ and under a wrong one by $-0.066$ (per seed $-0.124 / -0.041 / -0.026$ and $-0.118 / -0.057 / -0.023$): the gap is arm-invariant, and 96\% of it is already present when the header R receives is the right one. R therefore never exhibits a wrong header pushing it below the baseline; it exhibits a checkpoint that sits below the baseline whatever the header says, which is a property of training on uncorrelated pairs. M has the opposite structure --- its correct arm is $+0.117$ \textit{above} the baseline and only the wrong-header substitution, worth $-0.198$, carries it across. This contrast is text-side, which is the tokenization the control was trained and evaluated under; for the token form M's own correct arm sits at 0.376 against the baseline's 0.406 at the program unit and straddles it per distinct input (§4.4), so on that side the argument that M pays no header-training cost is not available and we do not make it. The consequence for the claim is stated in full at the end of this section: what the control isolates is the \textit{content sensitivity}, not the below-baseline level.

\begin{figure}[!tb]
\centering
\includegraphics[width=0.62\columnwidth]{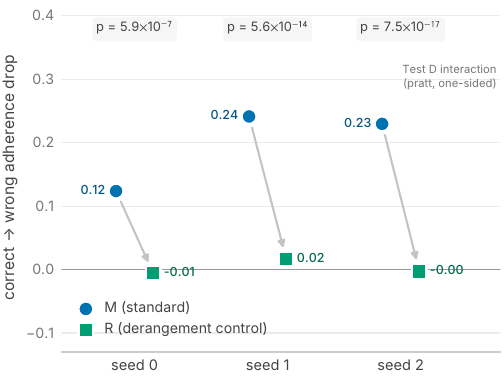}
\caption{\textbf{The dissociation, on the replication sample.} Correct$\rightarrow$wrong adherence drop per seed for the standard model M (circles) and the derangement control R (squares) over the 400-program held-out set; p-values are the per-seed Test D interaction (paired Wilcoxon, \texttt{pratt}, one-sided). M is consistently hurt by a wrong header; R --- trained on the identical header marginal but with no learnable header$\rightarrow$program mapping --- is not.}
\label{fig:the_dissociation_on_the_replic}
\end{figure}

\begin{table}[ht]
\centering
\caption{\textbf{Test D on the replication sample --- the primary panel. The earlier sample is reported alongside it, as the frozen retest protocol requires, in Table 4.} Both runs are reported, as the frozen retest protocol requires, with the replication run primary. All cells are text headers, 40\% prefix, three seeds; the test is the same-seed paired one-sided Wilcoxon (\texttt{pratt}, asymptotic) on the per-sample interaction drop$_M$ - drop$_R$, identical in both tables --- the retest protocol changed the analysis sample and nothing else. The two tables are not on a common scale and are never pooled: they differ in intent-profile composition (eleven stratified profiles against six) and in the generation-truncation regime of their held-out sets, so only within-table contrasts are interpretable. Arm-level absolute adherence is kept out of this table; the replication run's arm levels are reported in the surrounding text and in §4.3. Analysis set: 400 deduplicated unique held-out programs, eleven four-tag intent profiles, non-zero-pair guardrail 60; the reference panel's is given in Table 4. The report-only content-blindness probe is annotated in Appendix A.}
\label{tab:3}
\small
\begin{tabularx}{\columnwidth}{lrrrXX}
\toprule
\textbf{Seed} & \textbf{drop$_M$} & \textbf{drop$_R$} & \textbf{Interaction} & \textbf{$p$ (pratt, one-sided)} & \textbf{Non-zero pairs (of 400)} \\
\midrule
seed 0 & 0.124 & -0.006 & 0.130 & $5.9{\times}10^{-7}$ & 156 \\
seed 1 & 0.241 & +0.016 & 0.225 & $5.6{\times}10^{-14}$ & 171 \\
seed 2 & 0.230 & -0.003 & 0.233 & $7.5{\times}10^{-17}$ & 159 \\
\bottomrule
\end{tabularx}
\end{table}

\begin{table}[ht]
\centering
\caption{Test D on the earlier held-out sample (76 entries over 38 unique programs, six profiles, non-zero-pair guardrail 15) --- the reference run for Table 3. Reported alongside the replication run as the frozen retest protocol requires; the two are never pooled or placed on a common scale.}
\label{tab:4}
\small
\begin{tabularx}{\columnwidth}{lrrrXX}
\toprule
\textbf{Seed} & \textbf{drop$_M$} & \textbf{drop$_R$} & \textbf{Interaction} & \textbf{$p$ (pratt, one-sided)} & \textbf{Non-zero pairs (of 76)} \\
\midrule
seed 0 & 0.184 & -0.007 & 0.191 & $4.2{\times}10^{-3}$ & 26 \\
seed 1 & 0.289 & -0.007 & 0.296 & $6.7{\times}10^{-6}$ & 26 \\
seed 2 & 0.283 & -0.059 & 0.342 & $3.6{\times}10^{-6}$ & 32 \\
\bottomrule
\end{tabularx}
\end{table}

On the 400 programs M's correct$\rightarrow$wrong drop is 0.124 / 0.241 / 0.230 per seed while R's is $-0.006 / +0.016 / -0.003$, that is indistinguishable from zero and inconsistent in sign (Figure 2; Table 3). The per-sample interaction is positive and significant on 3/3 seeds ($p = 5.9{\times}10^{-7} / 5.6{\times}10^{-14} / 7.5{\times}10^{-17}$; means 0.130 / 0.225 / 0.233), with no reverse-direction veto and no seed flagged underpowered --- 156 / 171 / 159 non-zero pairs against the pre-registered guardrail of 60, where the earlier run rested on 26 / 26 / 32 of 76. No seed is marginal in the way the earlier run's seed 0 was: the weakest seed of the retest is significant at $5.9{\times}10^{-7}$, and the $\geq 2/3$ rule is satisfied by any two of the three. All three pre-registered guardrails hold. On the four-arm executable intersection, where execution-inclusive zero-filling is removed entirely, the interaction is larger rather than smaller, at $+0.274$ over $N = 467$ sample$\times$seed pairs (114 / 191 / 162 per seed), so the zero fill is the conservative direction and not the source of the effect. The profile-level direction guardrail is $+0.283$, with all ten counted profiles individually positive and the eleventh, reported but not counted, positive as well. The earlier run reached the same ruling under the identical rule on its own sample (Table 4).

\begin{figure}[!tb]
\centering
\includegraphics[width=0.62\columnwidth]{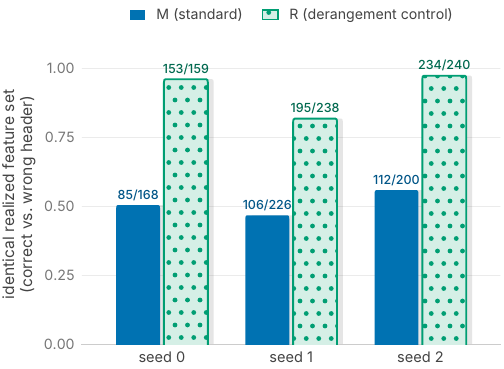}
\caption{\textbf{Output identity.} Fraction of jointly executable samples whose realized geometric feature set is \textit{identical} under correct and wrong headers, on the 400-program replication sample. The control R emits the same geometry regardless of header content on 82--98\% of samples; the standard model M does so on about half (47--56\%) --- its outputs change with the header's content. R's competence and its margins are reported in the text rather than assumed here.}
\label{fig:output_identity_fraction_of_jo}
\end{figure}

The most direct view of the dissociation does not run through the metric at all. On jointly executable outputs R realizes an identical geometric feature set under correct and wrong headers on 96\% / 82\% / 98\% of samples per seed (153/159, 195/238, 234/240; Figure 3), while M does so on 51\% / 47\% / 56\% (85/168, 106/226, 112/200). In this completion regime the control does not route header content into generation; the standard model does. The earlier sample gave the same picture on R denominators of 30--42 rather than 159--240.

That M's geometry moves is not by itself evidence that it moves \textit{toward} what the wrong header names. The quantity that settles this is a within-arm contrast, computed only on features the true intent does not contain: how often the model realizes such a feature when the injected header names it, against how often it realizes the same feature when the header does not. Because both terms come from the same arm, the comparison is not distorted by that arm's marginal rate of emitting features or by how hard its samples are. It is not immune to executability, and we state the direction of that dependence rather than assume it away: a program that does not run realizes nothing and so contributes alike to both branches, which attenuates the all-sample rows toward zero. Those rows are therefore conservative, but their magnitudes are not comparable across arms that run at different rates; the executable-only rows remove the attenuation, at the cost of conditioning on an outcome the arm affects --- and one the header moves differentially by feature, since whether a program runs itself depends on which features the header names. Both readings are therefore descriptive; neither is offered as a test. Table 5 gives it. On three of the four scored features the standard model moves substantially toward the false target under both tokenizations, and R --- handed byte-identical wrong headers --- shows no movement the data can detect, its largest entry being TALL at $+0.032$ with a 95\% interval covering zero. CIRCLE is the exception in every arm, and the reason is visible in §3.6: \texttt{.circle(} appears early in a program, CIRCLE dominates the prefix-leakage inventory, and a feature already fixed by the prefix is not available for a header to move.

\begin{table}[ht]
\centering
\caption{\textbf{c. Movement toward the injected wrong target (replication sample, 40\% prefix, pooled over three seeds; report-only).} For each scored feature $t$, the entry is $P(t \text{ realized} \mid t \in \text{injected header}) - P(t \text{ realized} \mid t \notin \text{injected header})$, computed only on samples whose \textit{true} intent does not contain $t$, so a positive value is uptake of a feature the program is not supposed to have. $p$ is a one-sided Fisher exact test pooling the three seeds, which treats the 1,200 rows of an arm as independent and so is anticonservative relative to the 298-distinct-input unit §4.1 states results at; nothing here is a decision and the entries are reported for their size rather than their $p$. The control R receives the identical injected headers at the identical indices and serves as the placebo row. Nothing here was pre-registered and no decision rests on it.}
\label{tab:5}
\small
\begin{tabularx}{\columnwidth}{XXXXXX}
\toprule
\textbf{Arm} & \textbf{Samples} & \textbf{CIRCLE} & \textbf{NGON} & \textbf{THIN} & \textbf{TALL} \\
\midrule
M, text wrong & all & $+0.016$ & $+0.325$ & $+0.313$ & $+0.347$ \\
M, text wrong & executable only & $+0.020$ & $+0.493$ & $+0.617$ & $+0.405$ \\
M, token wrong & all & $+0.009$ (n.s.) & $+0.168$ & $+0.073$ & $+0.330$ \\
M, token wrong & executable only & $+0.014$ (n.s.) & $+0.305$ & $+0.323$ & $+0.389$ \\
R, wrong (placebo) & all & $+0.005$ (n.s.) & $+0.009$ (n.s.) & $-0.025$ (n.s.) & $+0.032$ (n.s.) \\
R, wrong (placebo) & executable only & $+0.007$ (n.s.) & $+0.017$ (n.s.) & $-0.032$ (n.s.) & $+0.051$ (n.s.) \\
\bottomrule
\end{tabularx}
\end{table}

The three large entries in each row are significant at $p < 10^{-18}$; the two CIRCLE entries for the text arm are the weakest that still reach $\alpha$ ($p = 1.4{\times}10^{-2}$ and $2.7{\times}10^{-2}$), THIN on the token arm is at $6.0{\times}10^{-3}$, and every entry marked n.s. has $p > 0.07$. On executable outputs the two tokenizations are of comparable magnitude, and the per-seed values are consistent: NGON $+0.544 / +0.507 / +0.409$ and THIN $+0.578 / +0.655 / +0.612$ for text, NGON $+0.310 / +0.278 / +0.323$ and THIN $+0.238 / +0.357 / +0.369$ for token. Three checks separate this from artefacts of the measure. Re-assigning each header to a different evaluation item, 2,000 times, holds every arm's output marginal fixed while destroying the header-item correspondence, and the observed value sits 8--11 standard deviations above that null on 3/3 seeds for both tokenizations ($p_{\text{perm}} \leq 5{\times}10^{-4}$). On features that are in \textit{neither} the injected header nor the true intent the same arms score $-0.04$ to $-0.05$ relative to the baseline, so this is redistribution toward the named target rather than a general willingness to emit features. And the token arm emits fewer features per sample than the baseline overall while hitting the injected wrong ones more often, which the ``it simply emits more tags'' reading cannot accommodate.

Because R is impaired, its flat correct-to-wrong difference admits an alternative reading that must be answered rather than assumed away: a control with less adherence to begin with has less adherence to lose, so drop$_R \approx 0$ could reflect a compressed range rather than immunity. Three measurements separate the two. First, the range was not exhausted. R's correct-header adherence is 0.267--0.407 and the metric is bounded below by zero, so a fall the size of M's (0.124--0.241) was expressible on every seed. Second, the compression argument is strongest exactly where zero-filled non-executable outputs pile up at the bottom of the scale, and removing them strengthens the dissociation instead of weakening it: on the subset where all four arms produce executable geometry, the interaction is $+0.197$ ($N = 114$), $+0.316$ ($N = 191$) and $+0.278$ ($N = 162$) per seed, against execution-inclusive values of $+0.130 / +0.225 / +0.233$ on the same seeds. Scoring each arm on its own executable outputs locates R on the same scale: it stands at 0.628 (correct) and 0.621 (wrong) against the baseline's 0.673 and M's 0.836 correct / 0.510 wrong --- near the unconditioned baseline and far from the bottom of the scale, which is to say holding an adherence level it demonstrably could have lost, and losing 0.007 of it where M loses 0.326. Third, what R's outputs do under a wrong header is not to change slightly but not to change: a control compressed against a floor would still produce small non-zero drops on outputs that differ, whereas R's realized feature set is identical, not merely similar, on 82--98\% of jointly executable samples.

Two further checks bear on the mechanism the interaction implies. A wrong header does not derail generation: wrong-header executability is not depressed relative to the no-header baseline (0.638 against 0.603 pooled; 0.618 / 0.680 / 0.618 against 0.583 / 0.578 / 0.650, higher on two of three seeds), so wrong headers yield programs that run at least as often as unconditioned ones while adhering less to the true intent --- misdirection rather than information loss. The executable-intersection deficits of §4.1 point the same way on the metric side, but they condition on a post-treatment variable and are not load-bearing here. Affirmative evidence internal to M points the same way, from the earlier evaluation's 0\%-prefix column: with a correct header M reconstructs the intent at 0.776 adherence, while with the header masked the same model on two of three seeds still emits runnable programs (96--100\% executable) that realize zero intent, and on the third collapses to no executable output. That column rests on very few distinct prompts (§4.5) and is reported as a supporting observation only.

The dissociation is also not carried by one intent family. Re-running Test D on the replication sample with each intent profile's programs removed in turn leaves 3/3 direction and 3/3 significance for all eleven single-profile removals, including the 112 THIN programs and the 39 pure-NGON ones. Removing every profile that contains a given feature is the stricter form of the check: deleting all 149 circle-containing programs likewise leaves 3/3 direction and 3/3 significance, and deleting every one of the 160 NGON-containing programs leaves 3/3 direction with 2/3 significance (seed 0 $p = 0.060$), satisfying the pre-registered $\geq 2/3$ rule. On the earlier sample, removing NGON cost seed 0 its significance ($p = 0.32$); that dependence does not reproduce, and we read it as a small-sample property of 22 NGON entries. Full robustness --- five zero-handling treatments, a diff-mass decomposition, and the complete leave-one-profile-out and leave-one-feature-out tables --- is in Appendix B.

Taken together, a model that could not have learned the header$\rightarrow$program mapping is unaffected by whether the header it receives is right or wrong, while the model that learned the mapping is measurably hurt by the same substitution, on the same programs, with byte-identical wrong headers injected at the same indices. The model's sensitivity to header \textit{content} therefore requires the learned mapping, and the marginal/mechanical distribution-shift confound is excluded: the vulnerability is not a reaction to receiving an unusual input, since for R an uncorrelated header is not unusual at all.

We state the boundary of that conclusion as precisely as the design allows, because a stronger reading is available and is not supported. What is established is the content sensitivity: the correct-to-wrong change, which is zero for a model without the mapping and large for a model with it, and which resolves into movement toward the named false target (Table 5). What is \textit{not} established is that the wrong arm's position \textit{below the never-header-trained baseline} requires the mapping. Testing that would need a mapping-free control whose level is otherwise comparable, and R is not one: it sits below the baseline by the same margin under a correct header as under a wrong one, so its level carries the cost of derangement training rather than the cost of a wrong header, and the two cannot be separated within this design. A control trained with a represented unconditional mode, or one evaluated with an unconditioned arm of its own, would decide it; neither exists here. The cross-baseline ordering of §4.1 stands as a measured property of the system --- a header-trained model handed a wrong header does worse than a model never trained with one --- and this section explains the mechanism inside M rather than attributing that ordering to it. Both legs of the argument now rest on one sample --- the inferential leg of §4.1 and the causal leg here are computed over the same 400 programs, the same eleven-profile stratification and the same three seeds --- so the claim no longer bridges two differently composed evaluation sets, and the causal leg now rests on an order of magnitude more unique programs than the earlier run's (400 against 38), with 156--171 non-zero pairs per seed against that run's 26--32. What did not change is the grain of the conclusion. The derangement design equalizes the header marginal and nothing finer, so it does not separate graded semantic reading from learned consistency-checking, and R's competence margin on this sample is thinner than the protocol anticipated, so the control's validity rests on the pre-registered gate together with the floor-independent and output-identity evidence above rather than on a comfortable margin (§5, §6).

\subsection{Heterogeneity across intent profiles}

\label{sec:heterogeneity_across_intent_profiles}

An earlier version of this work scoped the wrong-vs-none claim to the design intents the model could render unconditioned --- polygonal and thin geometries --- on the ground that the pure-CIRCLE and pure-TALL profiles were floored at baseline, with adherence 0.00 in both compared arms. That restriction was a property of the sample rather than of the model. The earlier analysis subset covers 38 unique programs across six profiles, three of them pure-CIRCLE and two pure-TALL, and contains no programs at all in five of the eleven profiles the clean held-out pool presents; on three and two programs, a seed-averaged 0.00 in two arms is indistinguishable from an empty cell. With fifteen or more programs per profile the floors are not there: pure-CIRCLE has an unconditioned baseline of 0.475, second only to pure-THIN among the single-feature profiles, and pure-TALL has 0.156 with a correct-header arm of 0.978. We therefore do not re-scope the claim to a different set of profile names, which would repeat the same inference on the same kind of evidence, but state it at the level at which it was tested --- the profile-balanced sample as a whole --- and disclose the heterogeneity here. Tables 6 and 7 give the two contrasts the paper uses, each beside the arm levels it is computed from, over all eleven profiles. Nothing in this subsection is a test: no per-profile p-value or power was pre-registered or computed, several profiles rest on the fifteen-program stratification quota, and the two pre-registered guardrails act on column means, not on rows.

\begin{table}[ht]
\centering
\caption{\textbf{Per-profile wrong-vs-none contrast with its arm levels (replication sample, text headers, 40\% prefix, seed-averaged; descriptive).} The contrast is measured against each profile's \textit{unconditioned} baseline, which is why the arm levels are shown beside it. The guardrail mean is the pre-registered direction check over the ten profiles with $n \geq 15$; CIRCLE+NGON+TALL exhausts its pool at six programs and is reported but not counted.}
\label{tab:6}
\small
\begin{tabularx}{\columnwidth}{lrrrrX}
\toprule
\textbf{Profile} & \textbf{$n$} & \textbf{baseline} & \textbf{M correct} & \textbf{M wrong} & \textbf{wrong - baseline} \\
\midrule
THIN & 112 & 0.592 & 0.568 & 0.440 & -0.152 \\
NGON+THIN & 70 & 0.438 & 0.400 & 0.305 & -0.133 \\
CIRCLE+THIN & 58 & 0.279 & 0.284 & 0.233 & -0.046 \\
CIRCLE & 40 & 0.475 & 0.525 & 0.342 & -0.133 \\
NGON & 39 & 0.179 & 0.786 & 0.291 & +0.111 \\
CIRCLE+NGON & 15 & 0.133 & 0.367 & 0.189 & +0.056 \\
CIRCLE+\discretionary{}{}{}NGON+\discretionary{}{}{}THIN & 15 & 0.289 & 0.474 & 0.348 & +0.059 \\
CIRCLE+TALL & 15 & 0.489 & 0.533 & 0.389 & -0.100 \\
NGON+TALL & 15 & 0.311 & 0.744 & 0.233 & -0.078 \\
TALL & 15 & 0.156 & 0.978 & 0.111 & -0.044 \\
CIRCLE+\discretionary{}{}{}NGON+\discretionary{}{}{}TALL & 6 & 0.444 & 0.519 & 0.315 & -0.130 \\
\textbf{guardrail mean ($n \geq 15$)} &  &  &  &  & \textbf{-0.046} \\
\bottomrule
\end{tabularx}
\end{table}

\begin{table}[ht]
\centering
\caption{\textbf{Per-profile dissociation with its arm levels (replication sample, text headers, 40\% prefix, seed-averaged; descriptive).} The interaction is drop$_M$ - drop$_R$ and is measured against the \textit{correct-header} arm of each model, which is a different reference from Table 6. Every profile's point estimate is positive.}
\label{tab:7}
\small
\adjustbox{max width=\columnwidth}{%
\begin{tabular}{lrrrrrr}
\toprule
\textbf{Profile} & \textbf{$n$} & \textbf{M correct} & \textbf{M wrong} & \textbf{R correct} & \textbf{R wrong} & \textbf{Interaction} \\
\midrule
THIN & 112 & 0.568 & 0.440 & 0.446 & 0.438 & +0.119 \\
NGON+THIN & 70 & 0.400 & 0.305 & 0.357 & 0.331 & +0.069 \\
CIRCLE+THIN & 58 & 0.284 & 0.233 & 0.239 & 0.236 & +0.049 \\
CIRCLE & 40 & 0.525 & 0.342 & 0.475 & 0.483 & +0.192 \\
NGON & 39 & 0.786 & 0.291 & 0.188 & 0.205 & +0.513 \\
CIRCLE+NGON & 15 & 0.367 & 0.189 & 0.122 & 0.122 & +0.178 \\
CIRCLE+\discretionary{}{}{}NGON+\discretionary{}{}{}THIN & 15 & 0.474 & 0.348 & 0.222 & 0.267 & +0.170 \\
CIRCLE+TALL & 15 & 0.533 & 0.389 & 0.511 & 0.511 & +0.144 \\
NGON+TALL & 15 & 0.744 & 0.233 & 0.289 & 0.311 & +0.533 \\
TALL & 15 & 0.978 & 0.111 & 0.133 & 0.133 & +0.867 \\
CIRCLE+\discretionary{}{}{}NGON+\discretionary{}{}{}TALL & 6 & 0.519 & 0.315 & 0.463 & 0.463 & +0.204 \\
\textbf{guardrail mean ($n \geq 15$)} &  &  &  &  &  & \textbf{+0.283} \\
\bottomrule
\end{tabular}
}
\end{table}

The two columns the paper relies on point different ways on some profiles, and the arm levels show why. Eight of the eleven wrong-vs-none point estimates are negative and three positive, while all eleven interaction estimates are positive; the profiles on which the signs diverge are among those with the lowest unconditioned baselines. Pure NGON is the clearest case. Its baseline is 0.179 --- this checkpoint can barely produce a polygonal loop without a header --- so a wrong header still leaves generation above that low floor, and the pure-NGON point estimate for wrong-vs-none runs in the opposite direction from the aggregate, at $+0.111$ on 39 programs, even though the correct-to-wrong fall on the same programs, $0.786 \rightarrow 0.291$, is among the largest in the table. Pure TALL is the same arithmetic at its extreme: correct 0.978, wrong 0.111, baseline 0.156, giving a wrong-minus-baseline of only $-0.044$ alongside the table's largest interaction, $+0.867$. The two contrasts measure different quantities against different references and their divergence on these rows follows from the height of the baseline, not from an inconsistency in the evidence.

Read across the four single-feature profiles, the same arithmetic exposes a pattern in where conditioning does its work. Ordering by unconditioned baseline gives THIN 0.592, CIRCLE 0.475, NGON 0.179, TALL 0.156; ordering by the gain a correct header brings gives TALL $+0.822$, NGON $+0.607$, CIRCLE $+0.050$, THIN $-0.024$; and ordering by the Test D interaction gives TALL $+0.867$, NGON $+0.513$, CIRCLE $+0.192$, THIN $+0.119$. The two geometries the model is worst at producing unconditioned are the two the header helps most, and also the two on which the model's output depends most heavily on the header's content. As a description of these eleven profiles the pattern is internally coherent: conditioning carries the most of the generation exactly where the unconditioned model is weakest, and it is there that corrupting the signal has the most to displace. These are seed-averaged point estimates on 15 to 112 programs with no test attached, and we offer them as an account of the observed spread rather than as a finding about intent families.

What this subsection does not support is a per-family claim in either direction. It does not license the earlier scoping, which the sample has contradicted; and it does not license its inverse, a claim that the harm is a property of the intent distribution rather than of particular intent families, because the heterogeneity is strong and three of the eleven wrong-vs-none point estimates run the other way. Which intent families the harm is largest or smallest for is a question this design cannot answer, and the licensed statement remains the aggregate one of §4.1: on the profile-balanced population the stratified draw defines, a wrong header is worse than never conditioning, and that aggregate is not an artifact of which profiles the draw over-sampled.

\subsection{What the independent metric changes}

\label{sec:what_the_independent_metric_changes}

\begin{figure}[!tb]
\centering
\includegraphics[width=\columnwidth]{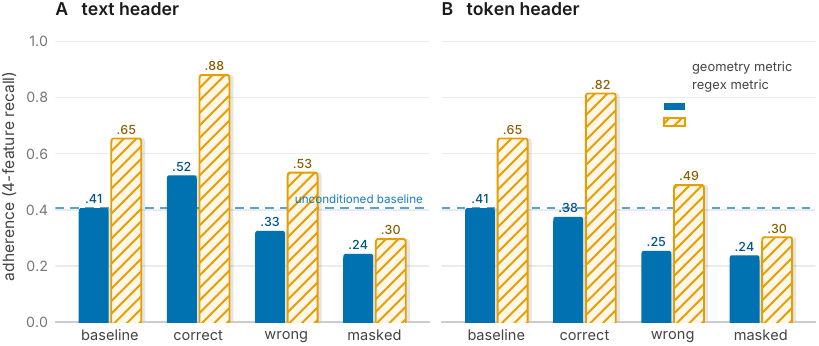}
\caption{\textbf{Adherence at 40\% prefix under the geometry metric (solid) against the training regex (hatched), per header arm; replication sample, mean over three seeds.} The dashed line marks the unconditioned baseline under the geometry metric. Both wrong-header arms fall below it --- the harm of §4.1 --- while the regex metric over-credits the \textit{correct}-header benefit, most starkly for the token header, whose correct arm does not clear the baseline under realized geometry at the program level and is indistinguishable from it per distinct input (§4.4). Wrong versus never-conditioned is stated for the tested sample as a whole, without restriction by design-intent family; descriptive per-profile detail is in §4.3 and Appendix C. Numbers in Table 8.}
\label{fig:adherence_at_40_prefix_under_t}
\end{figure}

\begin{table}[ht]
\centering
\caption{\textbf{Arm levels under the independent geometry metric, the training-side regex, and executability, with three set-valued alternatives to the primary metric (replication sample, 40\% prefix, mean over three seeds).} The first two metric columns use the same four-tag convention and are computed on the same generations, so the difference between them is the instrument and nothing else. The pre-registered decisions rest on the geometry column alone; the last three columns are report-only and answer a property of the primary metric rather than of the model. Adherence is a recall over the intended features and carries no penalty for realizing a feature the intent did not ask for (§3.4), so micro-averaged precision, micro-$F_1$ and exact-feature-set accuracy are given beside it. They leave the main ordering intact --- the correct arms are the most precise (0.974 text, 0.953 token) and the wrong arms the least (0.688, 0.754), and wrong stays below the unconditioned baseline on $F_1$ and on exact-set accuracy under both tokenizations --- with one exception worth stating: the wrong-above-masked ordering of §4.1 holds on $F_1$ for the text form (0.429 against 0.384) but not for the token form (0.371 against 0.378), where the two arms are separated only on exact-set accuracy (0.176 against 0.165). Precision behaves as the metric's blind spot predicts: the masked arms score high on it (0.876, 0.893) while realizing almost nothing.}
\label{tab:8}
\small
\adjustbox{max width=\columnwidth}{%
\normalsize\setlength{\tabcolsep}{4pt}
\begin{tabular}{lrrrrrr}
\toprule
\textbf{Arm} & \textbf{geometric recall} & \textbf{regex recall} & \textbf{executability} & \textbf{micro-P} & \textbf{micro-$F_1$} & \textbf{exact set} \\
\midrule
unconditioned baseline & 0.406 & 0.655 & 0.603 & 0.798 & 0.520 & 0.305 \\
text --- correct & 0.523 & 0.881 & 0.626 & 0.974 & 0.649 & 0.434 \\
text --- wrong & 0.325 & 0.533 & 0.638 & 0.688 & 0.429 & 0.198 \\
text --- masked & 0.243 & 0.296 & 0.855 & 0.876 & 0.384 & 0.167 \\
token --- correct & 0.376 & 0.815 & 0.477 & 0.953 & 0.505 & 0.302 \\
token --- wrong & 0.254 & 0.489 & 0.507 & 0.754 & 0.371 & 0.176 \\
token --- masked & 0.238 & 0.303 & 0.828 & 0.893 & 0.378 & 0.165 \\
control R --- correct & 0.343 & --- & 0.546 & 0.788 & 0.464 & 0.256 \\
control R --- wrong & 0.340 & --- & 0.548 & 0.789 & 0.462 & 0.251 \\
\bottomrule
\end{tabular}
}
\end{table}

The independent metric deflates the apparent benefit of conditioning, and does so more sharply on this sample than on the earlier one (Figure 4). Under the convention-matched four-tag comparison of Table 8, the correct-header gain over the unconditioned baseline falls from $+0.226$ under the training-side regex to $+0.117$ under realized geometry for the text header, and from $+0.161$ to $-0.031$ for the token header. That last figure is unit-dependent and we report both units, since §4.1 states the paper's results at the level of distinct model inputs: the token correct arm sits $0.031$ below the unconditioned baseline per program, and is indistinguishable from it per distinct input ($+0.008$, 95\% CI $[-0.024, +0.041]$; per seed $-0.001 / +0.025 / +0.000$). What the token form reliably differs in is not adherence but executability, as the decomposition below shows, so the honest summary is that the token header's benefit does not survive the change of instrument rather than that it reverses.

The two figures differ in two respects (§3.4), and the respects do not contribute equally. Inserting the intermediate quantity --- the training-side regex applied under the same execution-inclusive rule, so that a program which does not run scores zero on it too --- splits the fall into a part owed to requiring execution and a part owed to changing the detector: $+0.161 \rightarrow -0.014 \rightarrow -0.031$ for the token header and $+0.226 \rightarrow +0.125 \rightarrow +0.117$ for the text header. Requiring the program to run accounts for $-0.175$ of the token header's total fall of $-0.192$ and for $-0.101$ of the text header's $-0.109$; changing the detector accounts for the remaining $-0.017$ and $-0.008$. The inflation the training-side metric carries is therefore almost entirely execution blindness rather than disagreement between two detectors on the same runnable output, which is what the audit of the initial evaluation found and what this larger sample reproduces.

The decomposition leaves an executability result that the aggregate figure hides and that we state directly, because it is what the token header's shortfall is made of. Conditioning on the token form depresses how often the model produces a program that runs at all: 0.477 of its correct-arm generations execute, against 0.603 for the never-header-trained baseline and 0.626 for the text form (Table 8). The effect is a property of the surface form rather than of conditioning as such --- the same header content in a comment line leaves executability slightly above baseline --- and it is large enough to reverse the sign of the arm-level comparison on its own. Where both arms do produce runnable geometry the token header's benefit is positive and substantial: $+0.195$, $+0.194$ and $+0.188$ per seed over $N = 146 / 142 / 173$ paired samples ($p = 1.3{\times}10^{-8} / 2.2{\times}10^{-8} / 1.7{\times}10^{-9}$), with the text header at $+0.240 / +0.172 / +0.167$. The $-0.031$ is thus a statement about how often the token-conditioned model writes a program that runs, not about whether it follows the header when it does, and it is also the figure that does not survive the change of analysis unit. The per-seed pattern is consistent with the arm means: text correct-vs-baseline is positive on 3/3 seeds ($+0.052 / +0.185 / +0.115$) at the program level and on 3/3 at the cluster level ($+0.102 / +0.157 / +0.111$, pooled 95\% CI $[+0.084, +0.164]$), while token correct-vs-baseline is negative on 3/3 seeds per program ($-0.011 / -0.035 / -0.046$) and straddles zero per distinct input. All of these are reported descriptively, as no per-seed test of a correct-header benefit is pre-registered. The wrong-header deficit, by contrast, is preserved on both metrics and under both tokenizations: against the regex baseline of 0.655 the wrong arms score 0.533 (text) and 0.489 (token), and against the geometry baseline of 0.406 they score 0.325 and 0.254.

The same execution gate explains a discrepancy a reader would otherwise take at face value. Scored across all samples, the token wrong arm shows no aggregate movement toward the target its injected header names, while the text arm shows a large one. That gap is the gate rather than the model: a program that does not run realizes no features by construction, and the token arm produces such programs half the time, which removes exactly the samples on which uptake would be visible. Measured within the arm on features the true intent does not contain --- the comparison of Table 5, which is unaffected by how often the arm runs --- the token form moves toward the falsely named feature on the same three features, and on executable outputs at a comparable magnitude to the text form. Over all samples its THIN entry is the weakest of the six ($+0.073$), which is what the attenuation above predicts for the arm that runs least often.

The asymmetry between the two metric columns carries the consequence of this subsection. A conditioning benefit scored by the feature detector that defines the conditioning signal is partly an artifact of the instrument --- roughly half of the text-header gain and all of the token-header gain in this setting --- while the harm survives the change of instrument intact. A literature that evaluates ``the model follows the specification'' with the detector that wrote the specification will therefore over-report benefit while remaining blind to the failure mode of §4.1, since a wrong-signal arm is not part of such an evaluation in the first place. The two habits compound rather than cancel: the first inflates what conditioning is credited with, and the second removes the condition under which the credit would be checked. This is why the evaluation instrument is presented in §3.4 as method rather than offered here as a finding --- it is the reason the results above are believable, not a result in its own right.

\subsection{Regime specificity}

\label{sec:regime_specificity}

The harm is a property of the conditional-completion regime, and outside it the contrast is not merely absent but undefined. In the earlier evaluation's 0\%-prefix column the unconditioned baseline cannot generate valid CAD at all --- 0 of 100 outputs execute, adherence 0.000 --- while a wrong header still yields some executable, partially adhering programs (0.160 text / 0.077 token), so a wrong header scores \textit{above} a baseline that has nothing to score. ``Wrong < never conditioning'' presupposes that the never-header model can generate, and at 0\% prefix it cannot. The replication did not re-test that column, and the decision was frozen before any of its evaluation data existed. At 0\% prefix the prompt contains only the header, so the number of distinct model inputs is capped by header combinatorics rather than by the number of programs drawn: on the earlier sample the baseline and masked arms had exactly one distinct prompt each, and across 100 nominal samples the baseline arm produced a single unique generation and the masked arm one or two, while the header-bearing arms produced ten to sixteen; enlarging the draw from 100 to 400 programs raises those counts to one and at most nineteen, against 298 distinct inputs at 40\% prefix (§6). Re-running the column would have added cost without adding independent observations, which is why it was excluded ex ante rather than after inspecting results. The consequence for this paper is that the positive 0\%-prefix statements retained from the earlier evaluation rest on far fewer independent observations than their nominal $n$ suggests and are reported with that caveat, while the decision-critical statement in that regime --- that the ordering is undefined where the unconditioned model cannot generate --- does not depend on the count. The scope this establishes is not a peripheral one: conditional completion --- extending or editing a partially specified program --- is the workflow such models are being built to support, and it is the regime in which the harm is located.

\section{Discussion}

\label{sec:discussion}

The result that carries this paper is the negative one, and it is now causally grounded on the same programs that establish it. In the regime where a designer would actually use such a model --- completing or editing an existing partial program --- supplying a semantically wrong design-intent header leaves the system further from the true intent than never training the interface at all, and §4.2 shows that the model's response to the header's \textit{content} --- the whole of the correct-to-wrong change --- is a property of the learned header$\rightarrow$program mapping rather than of receiving an odd input. The control does not establish that the below-baseline level itself requires the mapping, and we do not claim it (§4.2). A system that trusts a possibly-wrong intent field will not merely be perturbed by it as noise; it will be \textbf{semantically misled} --- steered, competently and executably, away from the true design intent. The benefit side of conditioning, meanwhile, is smaller than its own self-referential evaluation reports: under the independent metric the correct-header gain is roughly halved for the text header, and for the token header it does not survive the change of instrument at all, in a direction that depends on the analysis unit (§4.4) --- almost entirely because the conditioned model more often writes a program that does not run, not because two detectors disagree about a program that does. Being explicit about that is the cost of scoring a produced solid rather than a produced string, and it is also what makes the negative result credible --- the same instrument that deflates the benefit preserves the harm.

The two models compared here differ in exactly one thing. They share a base checkpoint, an adapter configuration, a training corpus, a step count and a seed schedule; they were shown header strings that are byte-identical in form and in frequency; at evaluation they were given the same 400 programs, the same 40\% prefixes, and the same wrong headers injected at the same indices. The only difference is whether the training data contained a consistent relation between a header and the program that followed it. The model trained without that relation is, on this sample, unaffected by whether the header it receives is right or wrong: its adherence moves by -0.006, +0.016 and -0.003 across seeds, and on 82--98\% of jointly executable outputs it realizes a geometric feature set that is identical under the two conditions. It is not that this control is robust in any general sense --- its absolute competence is 0.34, below the unconditioned baseline's 0.406, and it is the weaker of the two models. It is that the header is not in its causal path at all. The control therefore supplies something a purely statistical argument cannot: an empirically measured zero point for what ``this architecture, this recipe, this header marginal, no learned mapping'' looks like when handed a wrong header. Against that measured zero, the standard model loses 0.124 to 0.241 of adherence and changes its realized geometry on roughly half of its jointly executable outputs. The significance of that asymmetry is specific. The vulnerability is not a defect that accompanied the conditioning mechanism; it is the conditioning mechanism, observed from the side on which the input is false. The learned header$\rightarrow$program mapping is what makes a correct header useful, and it is the same object that routes a wrong header into the geometry. A system builder therefore cannot treat capability and wrong-intent exposure as two properties to be traded against each other by tuning robustness, because in this setting they are not two properties: the model that reads the header is the model that can be misled by it, and the model that cannot be misled is the model that is not reading. This forecloses the two mitigations that would otherwise be reached for first. Input validation cannot help, because a wrong header is well-formed by construction --- it is another program's genuine header, and it passes any format, vocabulary or schema check that the correct one passes. And the comforting deployment assumption that a bad conditioning signal degrades a model toward its unconditioned behavior is not merely unsupported here but contradicted by direct measurement: if degradation were mechanical, a reaction to receiving an odd or unexpected input, the derangement control would show it too, since for that control an uncorrelated header is the training distribution itself. It shows none. What remains is that the failure has to be intercepted where the mapping is used --- at training time, by giving the model a represented option of proceeding unconditioned, or at decode time, by backing off toward an unconditioned stream when the conditioning signal is suspect. The masked arm shows why the unconditional mode has to be trained in rather than improvised: suppressing the header of a header-trained model at inference is itself an off-distribution input, and the arm it produces, while the most executable in the study, realizes no intent feature at all on 59\% of the programs that run (§4.1). A backoff target has to be a mode the model represents, not the absence of a field it expects. That is a narrower and more demanding target than input hygiene, and locating it is the practical yield of the control. Two limits on this reading should be kept in view. The design separates ``the header's content is in the causal path'' from ``the header's presence perturbs generation''; it does not separate graded semantic reading from learned consistency-checking, and the dose-response probe intended to do so did not confirm. And the control is impaired relative to the competence the protocol anticipated (§6), so the contrast is between the presence and absence of a header$\rightarrow$output causal path, not between two models of matched ability.

The derangement control equalizes the header \textit{marginal}, so it excludes explanations in which a wrong header hurts merely by being a rare, malformed, or length-shifted input --- the marginal/mechanical confound. It cannot exclude a subtler reading: for M, trained only on consistent header/program pairs, a wrong header creates a \textit{jointly} atypical (header, prefix) input, so one may posit that M detects the inconsistency and degrades, rather than ``believing'' the wrong content. We do not separate graded semantic reading from learned consistency-checking here, and the pre-registered dose-response probe that might have (§4.2) did not confirm. Re-running the control on the larger, profile-stratified sample does not touch this boundary: it raises the confidence that the content sensitivity requires the learned mapping, and shows the dissociation is not carried by one intent family, but the derangement design equalizes the header marginal and nothing finer, so the grain of the question --- what the model does with the content once it has represented it --- is exactly where it was. We note only that detecting a header/prefix inconsistency itself requires representing the header's content, so either reading places header \textit{content}, not input oddity, in the causal path; and the observed signature --- executability not depressed, adherence below baseline --- is that of misdirection rather than derailment.

This is the natural-language distractor phenomenon \cite{tao2026nwcad} instantiated in structured geometric program generation, and it motivates the same remedy: a conditioning interface that can fall back to unconditioned behavior when intent is absent or suspect. The header-dropout variant trained here is one cheap training-side step in that direction, and a decode-time backoff in the spirit of \cite{tao2026nwcad} is a natural next one; neither has been evaluated under the geometry metric, and the variant is described in Appendix D with the rest of the material that supports no claim in the main text.

\section{Limitations}

\label{sec:limitations}

The wrong<none effect holds in the 40\%-prefix conditional-completion regime, not the 0\%-prefix from-scratch regime, and the 40\% setting couples the effect to information carried by the program prefix. On the stratified 400-program sample, 17.8\% of programs have at least one four-tag intent feature already satisfied inside the prefix (18.0\% counting MULTI\_PART), materially more than on the earlier unstratified sample, where the figure was 8.0\%. The increase is a foreseeable consequence of the draw, which introduces CIRCLE-family profiles the earlier sample largely lacked, and \texttt{.circle(} appears early in a program. The leakage is symmetric --- every arm at a given index sees the identical prefix --- so it cancels in the paired differences that carry every decision, but it raises the recall floor and converts samples into exact ties, costing power. We do not claim a prefix-independent effect.

The 0\%-prefix regime was not re-tested, and the reason is structural. At 0\% prefix the prompt contains only the header, so the number of distinct inputs is capped by header combinatorics --- at most nineteen distinct prompts across that column, and exactly one for the baseline and masked arms --- independently of how many programs are drawn. On the earlier sample this was measurable on the output side: across 100 nominal samples the baseline arm produced a single unique generation and the masked arm one or two, while the correct and shuffled arms produced ten to sixteen; on the correct-header arm the cluster-level standard errors are 2.80--3.45$\times$ the naive ones. Re-running that column on 400 programs would add cost without adding information, which is why it was excluded ex ante rather than after seeing results. The consequence for this paper is that the positive 0\%-prefix statements retained from the earlier evaluation rest on far fewer independent observations than their nominal $n$ suggests, and are reported with that caveat; the decision-critical statement in that regime --- that ``wrong < never conditioning'' is undefined where ``none'' cannot generate at all --- does not depend on the count.

Three mechanisms reduce effective sample size on this sample, and all were recorded before the decision criteria were computed. The first two are symmetric across arms and cost power; the third is not a power problem at all, and we treat it separately below. First, the generation extractor truncates at the first top-level \texttt{result =} assignment, while a multi-part program's 40\% prefix can already contain a complete assignment block; on 22.1\% of scored rows the program that gets scored therefore lies entirely inside the prompt prefix and the model's own generation is discarded --- on the 72 indices where all three arms are affected, the paired difference is identically zero. That rule also introduces an asymmetry between the two sides of the metric, which we state because it changes the task and not only the power. The scoring \textit{target} is $\phi$ of the untruncated ground-truth program, while the scored \textit{output} is the generation cut at its first top-level \texttt{result =}; a model that faithfully continues into a second modelling stage is therefore not credited for it. Applying the same truncation to the 400 reference programs changes the four-tag target on 79 of them, 70 by losing a CIRCLE that appears only after the first block. Excluding those 79 leaves the effect at $-0.098 / -0.061 / -0.132$ text-side and $-0.182 / -0.145 / -0.193$ token-side, significant on 3/3 seeds in both at the program level --- but that comparison carries the composition caveat recorded below, and at the cluster unit the truncation exclusion alone leaves the text seed-1 cell at $p = 0.101$ (Appendix B), and the exclusion is not a substitute for the fix. Repairing the extractor changes a frozen component of the pipeline, so re-running under it is a separate, independently pre-registered experiment rather than an amendment to this one. That experiment has since been run --- pure inference on the checkpoints we retain, under criteria, an outcome mapping and the verbatim wording of all three possible outcomes fixed and hash-registered before the first generation job started. Repairing the extractor and re-running reproduces the wrong-versus-never-conditioning deficit on a scoring surface where the model's own continuation past the first \texttt{result} block is credited, at the cluster unit and under both tokenizations, with the same text seed failing there as fails here. The deficit reported above is therefore not an artifact of the truncation asymmetry. What that reproduction is worth is bounded by how much the repair changed: on 62.7\% of multi-stage generations the execution-driven rule returns exactly what the first-block rule returns, because the later modelling stages the model writes are themselves usually not executable, and the two surfaces differ on 11.2\% of scored rows. No figure from that protocol enters this paper; its pre-registration, verdict and run log are released with this version. Second, the 17.8\% prefix leakage above. Third, the 400 programs yield only 298 distinct 40\% prefixes, with a largest cluster of 41 programs sharing one input, so the header-free arms see identical inputs for those. The pre-registered non-zero-pair guardrail was set at 60 for exactly this reason; the observed counts were 113--126 for the primary cells and 156--171 for Test D, and no seed was flagged underpowered.

The third mechanism needs more than a count, and it is why §4.1 states the result at the level of distinct model inputs rather than of programs. Programs that share a prefix are not independent draws, and non-independence does not reduce power: it invalidates the null variance the paired test assumes, so a $p$ computed over 400 nominally independent pairs is anti-conservative. The frozen test specification re-run unchanged on the 298 distinct inputs is reported in §4.1 and is the primary reading; Test D's interaction is untouched by it ($p = 3.6{\times}10^{-9} / 1.9{\times}10^{-10} / 6.6{\times}10^{-10}$), five of the six primary cells survive, and the sixth --- text on seed 1 --- moves from $p = 0.011$ to $p = 0.098$ and is not called significant. It was an oversight in the earlier version of this work to have reported the clustering as a power cost and left the re-analysis in the limitations. The same re-analysis applied to the executable-intersection view of the same contrast leaves all six cells significant ($p \leq 1.5{\times}10^{-2}$), which is consistent with the failing cell being the one whose evidence rests most heavily on zero-filled non-executing programs; we record the observation but do not treat it as reinstating the cell, since that subset conditions on a post-treatment variable (§4.1). We report this rather than the nominal figure alone because the clustering was disclosed before the criteria were computed and the analysis costs nothing but a re-run: the same reason the earlier 0\%-prefix column was given cluster-level standard errors (§4.5) applies here, and it was an oversight not to have applied it to the column that carries the claims.

The aggregate harm is not an artifact of which profiles the stratified draw over-samples --- the report-only reweighting to natural held-out frequencies strengthens it --- but the per-profile point estimates are strongly heterogeneous and three of eleven run positive for the wrong-vs-none contrast. No per-profile test, p-value or power is claimed, and no claim in this paper is attached to an individual intent family; which families the harm is largest or smallest for is an open question this design cannot answer. The primary metric remains a four-feature, three-valued mean per-sample positive-label recall, a coarse instrument that rewards realizing the intended features and does not penalize realizing others; MULTI\_PART is excluded as geometrically under-determined.

A share of the reference corpus is geometrically degenerate, which caps what the metric can register and which we had not previously measured. Of the 400 frozen ground-truth programs, 62 contain a zero-depth \texttt{.extrude(0.0)} and none of those 62 executes; 129 in total fail to execute, and ground truth scores 0.567 against its own four-tag target (§3.4). Two further properties of the treatment are worth stating plainly. Because the header carries five features while the metric scores four, a wrong header can differ from the correct one on MULTI\_PART alone: 29, 22 and 33 of the 400 injected wrong headers per seed are identical to the correct header in the four scored tags, and are null treatments in the measured space. And the 202 programs whose header carries MULTI\_PART inherit a tag that only 34 of them realize as more than one solid.

Excluding the degenerate ground truth, the null treatments and the truncation-affected targets makes the effect larger --- text $-0.125 / -0.086 / -0.157$ and token $-0.220 / -0.156 / -0.222$, significant on 3/3 seeds in both --- and we report that without presenting it as evidence of robustness, because most of the increase is compositional rather than substantive. Eighty-four to eighty-six per cent of samples sit at a baseline recall of exactly 0 or exactly 1, where the paired difference is sign-constrained by the bound of the metric; every one of these exclusions enriches the upper stratum. Standardizing the retained samples' within-stratum effects back to the full sample's baseline composition returns $-0.090 / -0.058 / -0.123$ against the unexcluded $-0.072 / -0.056 / -0.115$, and a content-blind placebo filter that removes the same number of samples matched on the baseline-recall distribution reproduces 74\%, 83\% and 76\% of the apparent gain. The exclusions therefore do not overturn the effect --- its sign is negative in all six cells under every combination we ran, and the pre-registered rule is met throughout --- but their magnitude is not informative about it. The null-treatment exclusion additionally changes the estimand, from the effect of a wrong header to the effect of a header wrong in the tags the metric scores. Appendix B gives the full lattice, including the eight-specification range for the one cell that fails at the cluster level, whose $p$ runs from 0.018 to 0.101 depending on which exclusions are applied; that cell is reported as failing and no combination is presented as rescuing it.

The control is weaker than its expected signature. The frozen protocol chose the 0.30 competence floor expecting R to land near the unconditioned baseline, around 0.42. It did not: R's per-seed adherence is 0.267 / 0.354 / 0.407 under correct headers and 0.273 / 0.338 / 0.410 under wrong ones, so seed 0 falls below the floor in \textit{both} arms and the gate passes only under the pre-registered $\geq$2/3-seed rule. R's executability clears the margin on the pooled figure the gate is defined on --- 0.547 against the baseline's 0.603, threshold 0.503 --- while on seed 0 alone R sits 0.17 below baseline (0.417 against 0.583). On executable outputs only, pooled over its two arms, R's adherence is 0.624 against the baseline's 0.673 (Mann--Whitney two-sided, $p = 0.015$), where the earlier evaluation showed no detectable difference. R is therefore an impaired control rather than a floored one, and the dissociation must be read as a contrast between a model with a header$\rightarrow$output causal path and one without, not between two models of matched ability. R was also evaluated with correct and wrong headers only and has no unconditioned arm of its own, and its shortfall against the never-header-trained baseline is the same under a correct header as under a wrong one. Whether the \textit{level} of a wrong arm below that baseline requires the learned mapping is accordingly not testable within this design, and §4.2 states the causal claim at the content sensitivity the interaction does test. A control carrying a represented unconditional mode, or one given an unconditioned arm, would decide it.

A control with less headroom also has less adherence to lose, which is an alternative reading of its flat correct-to-wrong difference, and we state it rather than let the passing verdict stand alone. The executable-only view is the check that speaks to it, and it does not support the reading: on its own executable outputs R reaches 0.628 under a correct header and 0.621 under a wrong one, sitting near the unconditioned baseline's 0.673 rather than at a floor, so there is adherence available to lose and R does not lose it, while M loses 0.326 over the identical comparison (0.836 $\rightarrow$ 0.510). Consistently, the interaction grows rather than shrinks when the execution-inclusive zero fill is removed entirely: +0.274 on the four-arm executable intersection (per seed +0.197 / +0.316 / +0.278 on $N$ = 114 / 191 / 162). The derangement additionally leaves 8.8--9.7\% of training headers correct by chance, a residual that biases R toward content-sensitivity and makes its observed immunity conservative.

The two evaluations are not on a common scale: the earlier and the present evaluation differ in intent-profile composition and in generation-truncation regime, so their absolute per-arm levels are reported separately and are not compared cell by cell. This is a condition of the frozen protocol, not a post-hoc reading of the results, and it is not a claim that either evaluation sits systematically above or below the other; the direction of the difference varies by arm.

The completion cap is a further limit of the same kind. It was set at 474 tokens from the 95th percentile of twenty training programs' completions (§3.6), 21.0\% of generations on this sample reach it against 2.3\% of ground-truth continuations that would exceed it, and because the raw completions were not stored (§A.6) the per-arm cap-hit rates cannot be recovered after the fact. Whether the arms differ in how often they are truncated is therefore unknown; a 1$\times$/2$\times$ cap comparison and a per-sample cap-hit field are on the list for the next run, together with the raw generation logs that would make the question answerable at all.

What the header is, and is not, bounds the scope more than the model or the corpus does. CIRCLE, NGON, THIN and TALL are terminal-state properties of a produced solid, not the objects a CAD engineer means by design intent: not constraints, not parametric relations, not editable behaviour, not function. The header is also oracle-derived --- it is read off the answer by $\phi$ rather than supplied by a user or inferred by an upstream model --- which is what makes ``wrong'' well defined here and is also what limits the reading. If such a field were a user's editing instruction, a header differing from the original continuation would often be a legitimate counterfactual request rather than an error, and the wrong-header arm would be measuring something else entirely; the correctness axis is well posed only where the field is an estimate of an existing intent, as it is when an upstream module infers it. CADCON is therefore a controlled coarse-geometric-attribute proxy for a design-intent interface, and the results are about that proxy under one model, one corpus, one prefix fraction and greedy decoding.

A natural-language-prompt baseline remains future work. The evidence also comes from a single 1.5B model on DeepCAD; whether the deflation of the conditioning benefit and the wrong<none harm persist at larger scale and on other program representations --- native DeepCAD token sequences, richer B-rep or code formats --- is open. Both held-out sets are internal draws from the same corpus, and adopting a public executable benchmark (e.g., the CADTests/CADBench families) remains the planned strengthening.

One discrepancy in the reported numbers should be recorded explicitly. The regex-side numbers reported from the earlier evaluation for the correct, masked and baseline cells at 40\% prefix differ from that run's original values by up to six samples per cell in validity rate and by at most two in adherence recall ($|\Delta| \leq 0.0233 = 2/86$), because their reference values came from a different historical evaluation code path; the decision-critical wrong-header cells reproduce exactly on adherence recall, the quantity every decision rests on ($\Delta = 0$ on all six seed-cells), with a single one-sample validity difference (0.51 $\rightarrow$ 0.52) on the token cell at seed 2, and every geometry-side decision in this paper is computed within a single internally consistent run. The control pipeline's built-in check that M's arm means match the published values re-reads the same frozen artifacts and is an anti-drift check, not an independent reproduction.

\section{Reproducibility and release}

\label{sec:reproducibility_and_release}

Every protocol under which a number in this paper was produced was pre-registered: decision rules, statistical tests, thresholds and analysis scripts were frozen, with SHA-256 hashes recorded, \textit{before} the corresponding evaluation data existed, and no rule was changed afterwards. The freezing happened in two chains. The first covers the initial evaluation: the six-cell protocol was frozen before any scoring, and the control protocol with its analysis script (\texttt{f4\_analysis.py}) was frozen on 2026-07-07, the day before the control model's evaluations ran. The second covers the evaluation that carries every headline claim here: the replication protocol, its frozen 400-program stratified sample and its analysis script were frozen before the replication evaluations ran, together with a blinded power check executed before any effect direction or p-value was computed --- its script was AST-inspected to contain no mean, direction or Wilcoxon path, its threshold of 60 non-zero pairs (0.15$n$) was cleared with roughly twofold margin in every cell, and the protocol forbids enlarging the sample after it returns; and the Test D increment, which changes the analysis sample and nothing else, was frozen while the control had produced no data on that sample. That last freeze was asymmetric and we state which side was open: M's arms on the replication sample were already part of the primary result and had been inspected, while the control's correct-to-wrong drop was unobserved, so the value of the interaction was not knowable at freezing time (§3.6). Appendix A records the full chain, including one withdrawn launch of the retest and the verification that the already-frozen artifacts were unchanged by it.

We release the frozen analysis scripts with their hash manifest, both frozen held-out lists --- the initial 100-entry list with its 76-entry / 38-program four-feature analysis subset and six profile definitions, and the 400-program replication list with its eleven profile definitions and per-profile draw quotas --- the complete raw per-sample geometry scores (\texttt{geom\_scores.jsonl} for all arms and seeds of the initial, replication and retest evaluations), the concatenated analysis input from which the retest verdict was computed, the four verdict files, the blinded power check with its output, and the figure and table scripts that recompute every number in this paper, at https://github.com/Jacky628/cadcon-derangement-control. Two levels of reproducibility should be distinguished, because the release now supports one of them completely and the other up to its last step. Every number in this paper can be recomputed from the released per-sample scores with the released scripts, and the geometry scoring itself is robust to the environment: the runs used CadQuery 2.7.0 on OCP 7.8.1.1, and re-executing 400 of the frozen generations, drawn across all arms and seeds, on a machine with a freshly installed CadQuery 2.8.0 / OCP 7.9.3 and no attempt to match the original environment reproduces the recorded executability bit and the realized feature set on 400 of 400. Re-generating those outputs is now supported up to the checkpoints: the release carries the training and generation harness byte-identical to the file the campaign ran, and with it the DeepCAD-to-CadQuery transpiler and the extractor $\phi$ as the functions that were actually executed rather than as re-implementations; the 400 held-out programs materialized in full, each verified against the frozen sample list by hash; and the complete package versions of the interpreter the runs used. Two things remain outside it. The LoRA adapters are held until acceptance, so the last step from a trained checkpoint to a generation cannot yet be repeated by a reader. And the raw completions were never stored (§A.6): the recorded field is the output of one pass of the extractor, so EOS behaviour, completion-cap hits per arm, and re-extraction under a different rule are unrecoverable for this campaign. That is the one gap that cannot be closed retrospectively, and it is why future runs will store raw token IDs, raw text and EOS/cap flags. Numerical recomputation is therefore supported end to end; regeneration is not, and what the next release round adds is the adapters, and after that the only irreducible loss is the raw generation log. The library versions were never part of the gap: they are recorded in the frozen pre-registration and the interpreter the campaign ran under is retained. All experiments ran on a single consumer workstation with two RTX 3090 GPUs (24 GB each): the four evaluation suites are pure inference (42, 18, 21 and 6 jobs), each LoRA fine-tuning run (M, the unconditioned baseline, and the R variants; three seeds each) completes in single-digit GPU-hours, and the control's re-evaluation on the replication sample required no additional training, reusing the three derangement-trained checkpoints. LoRA adapter checkpoints for all reported conditions will be released upon acceptance.

\section{Conclusion}

\label{sec:conclusion}

This paper built a controlled diagnostic around a conditioned generator assembled in the pattern of current fine-tuning practice, conditioned on a controlled proxy for design intent rather than on design intent as a CAD engineer would state it: CADCON, a five-feature design-intent header prepended to CadQuery-style sketch-extrude programs and fine-tuned into a 1.5B code LLM on consumer hardware (model M, §3.3). Two instruments carry the diagnosis. The first is an execution-level adherence metric that scores the produced B-rep solid and shares no code with the extractor that defines the headers (§3.4), removing the produced-side circularity of self-referential evaluation. The second is a derangement-trained negative control, R --- identical training recipe, identical header marginal, no consistent header$\leftrightarrow$program relationship to learn (§3.5) --- which turns ``does the model read the header?'' into a testable contrast. Both were applied to one frozen sample of 400 deduplicated held-out programs, stratified over the eleven four-tag intent profiles the clean pool contains, so that the harm and the control that explains it are measured on the same programs, at the same indices, with byte-identical wrong headers. Every decision rule was pre-registered and frozen before the data that could decide it existed, in both freeze chains --- including the one asymmetric freeze recorded above --- and a blinded power check ran before any direction or p-value was computed.

Five results emerge, each computed under a frozen, pre-registered protocol. (i) Under the independent metric, a semantically wrong header degrades adherence \textit{below} the no-header baseline in conditional completion, direction-unanimous across three seeds and both tokenizations, and stated at the aggregate level over a profile-balanced population rather than scoped to a subset of intent families: a pre-registered replication on 400 deduplicated held-out programs stratified over all eleven observed intent profiles reproduces the deficit at 0.406 $\rightarrow$ 0.325 text / 0.254 token on that sample, significant on 3/3 seeds on both sides at the program level and on 3/3 token and 2/3 text seeds at the 298 distinct model inputs those programs present, with the pre-registered profile-level direction guardrail holding. (ii) The harm is causal in the precise sense the control can test (text headers): on the same 400 held-out programs that establish the harm, M's correct$\rightarrow$wrong drop (0.12--0.24 per seed) dissociates from R's ($\approx 0$) on 3/3 seeds ($p \le 5.9{\times}10^{-7}$), and the dissociation survives removal of any single intent profile --- the model's sensitivity to whether the header is right or wrong requires the learned header$\rightarrow$program mapping, and the marginal/mechanical distribution-shift confound is excluded; the control sits below the unconditioned baseline by the same margin under a correct header as under a wrong one, so it bounds the claim to that sensitivity and not to the below-baseline level itself. (iii) The harm is strongly heterogeneous across intent families while its aggregate does not depend on the stratified draw: the per-profile wrong-vs-none point estimates run from -0.152 to +0.111 with three of eleven positive, all descriptive and carrying no test, and a report-only reweighting to the profiles' natural held-out frequencies strengthens the aggregate to -0.086 against the guardrail's -0.046. The inferential population is accordingly a profile-balanced one, and no claim is attached to an individual intent family. (iv) The effect belongs to conditional completion: with no program prefix the unconditioned baseline cannot generate valid CAD at all, so ``wrong < never conditioning'' is undefined there, and that column was not re-tested because its effective sample size is capped by header combinatorics at nineteen distinct prompts however many programs are drawn (§6). (v) The apparent benefit of a \textit{correct} header is partly an artifact of the instrument that defines the conditioning signal: on the same 400 programs and the same generations, a convention-matched comparison takes the text header's correct-arm gain from +0.226 under the training-side regex to +0.117 under realized geometry, and the token header's from +0.161 to -0.031 per program --- a figure that is unit-dependent, since per distinct model input the token correct arm is indistinguishable from the baseline (+0.008, 95\% CI [-0.024, +0.041]). Decomposed, that fall is almost all the cost of requiring the program to run (-0.175 of the token header's -0.192) rather than of changing the detector (-0.017): what the training-side metric hides is chiefly that the conditioned model writes programs which do not execute, and the token form does so far more often than the text form. The wrong-header deficit is preserved on both metrics.

The two evaluation habits this diagnosis interrogates --- scoring conditioning with a signal-derived metric, and evaluating only correct signals --- each concealed something: the first inflated the benefit, the second hid an active hazard. Correcting both re-centers the headline: the durable result is not the benefit of a correct header --- real on the text side, absent for the token header once adherence is read off the produced geometry --- but that a model which has learned to rely on an intent field is actively misdirected by that field when it is wrong, producing competent, executable geometry that adheres to the true intent \textit{less} than a model never trained with the field --- on a profile-balanced held-out sample spanning every intent profile the corpus presents, on the workflow such models are built to support. To our knowledge this is the first causally-controlled isolation of wrong-content sensitivity in structured geometric program generation, and it locates the failure where repair can act: in the learned conditioning interface, not in input hygiene.

For LLM-based parametric CAD generation specifically, three consequences follow. For builders: a conditioning field --- a text description, a structured header, an upstream model's inferred plan --- must be treated as untrusted input, with the fallback to unconditioned behavior built into the training interface, not into post-hoc filtering, because a wrong header is well-formed and no filter distinguishes it from a right one. For evaluators: a conditioning claim should answer three questions before it is believed --- does the benefit survive a conditioning-independent, execution-level metric; is a wrong signal worse than never conditioning at all; and does the model's response to the signal's content require the learned mapping? This audit pair answers all three on a consumer workstation: the evaluations are pure inference, the derangement control adds one fine-tuning run per seed at single-digit GPU-hours, and re-testing that control on a new sample adds inference alone. For the field's trajectory: whether the deflated benefit and the wrong-header harm survive larger models, other program representations, and richer intent languages is now a sharp, testable question, as is which intent families carry the harm --- a question this design deliberately does not answer. Intent conditioning for CAD is worth building; interfaces that degrade gracefully to unconditioned behavior when intent is absent or wrong are worth building first.

\bibliographystyle{plainnat}
\bibliography{references}

\begin{thebibliography}{26}
\providecommand{\natexlab}[1]{#1}
\providecommand{\url}[1]{\texttt{#1}}
\expandafter\ifx\csname urlstyle\endcsname\relax
  \providecommand{\doi}[1]{doi: #1}\else
  \providecommand{\doi}{doi: \begingroup \urlstyle{rm}\Url}\fi

\bibitem[Casey et~al.(2025)Casey, Zhang, Ishida, Thompson, Khasahmadi,
  Lambourne, Jayaraman, and Willis]{casey2025aligning}
Evan Casey, Tianyu Zhang, Shu Ishida, John~Roger Thompson, Amir Khasahmadi,
  Joseph~George Lambourne, Pradeep~Kumar Jayaraman, and Karl D.~D. Willis.
\newblock Aligning constraint generation with design intent in parametric
  {CAD}.
\newblock In \emph{Proceedings of the IEEE/CVF International Conference on
  Computer Vision (ICCV)}, pages 8613--8622, 2025.
\newblock URL
  \url{https://openaccess.thecvf.com/content/ICCV2025/html/Casey_Aligning_Constraint_Generation_with_Design_Intent_in_Parametric_CAD_ICCV_2025_paper.html}.

\bibitem[Dettmers et~al.(2023)Dettmers, Pagnoni, Holtzman, and
  Zettlemoyer]{dettmers2023qlora}
Tim Dettmers, Artidoro Pagnoni, Ari Holtzman, and Luke Zettlemoyer.
\newblock Qlora: Efficient finetuning of quantized llms.
\newblock In \emph{Neural Information Processing Systems}, 2023.
\newblock \doi{10.48550/arXiv.2305.14314}.
\newblock URL
  \url{https://www.semanticscholar.org/paper/32ac52069e562d4f900afee70bdca63f53461481}.

\bibitem[Gong et~al.(2026)Gong, Wu, Liu, and Tu]{gong2026toolcad}
Yifei Gong, Xing Wu, Wenda Liu, and Kang Tu.
\newblock Toolcad: Exploring tool-using large language models in text-to-cad
  generation with reinforcement learning.
\newblock \emph{arXiv preprint arXiv:2604.07960}, 2026.
\newblock URL \url{https://arxiv.org/abs/2604.07960}.

\bibitem[Govindarajan et~al.(2025)Govindarajan, Baldelli, Pathak, Fournier, and
  Chandar]{govindarajan2025cadmium}
Prashant Govindarajan, Davide Baldelli, Jay Pathak, Quentin Fournier, and
  Sarath Chandar.
\newblock Cadmium: Fine-tuning code language models for text-driven sequential
  cad design.
\newblock \emph{Trans. Mach. Learn. Res.}, 2025.
\newblock \doi{10.48550/arXiv.2507.09792}.
\newblock URL
  \url{https://www.semanticscholar.org/paper/eb088897b07b26473068534bc6d31e3a6ad48696}.

\bibitem[Guan et~al.(2025)Guan, Wang, Ming, Zhang, Xu, and
  Yu]{guan2025cadcoder}
Y.~Guan, Xilin Wang, Xingxia Ming, Jing Zhang, Dong Xu, and Qian Yu.
\newblock Cad-coder: Text-to-cad generation with chain-of-thought and geometric
  reward.
\newblock \emph{arXiv.org}, 2025.
\newblock \doi{10.48550/arXiv.2505.19713}.
\newblock URL
  \url{https://www.semanticscholar.org/paper/dfd5941b9f60a97de58b6bc4da56c0a22b348580}.

\bibitem[Ho and Salimans(2022)]{ho2022cfg}
Jonathan Ho and Tim Salimans.
\newblock Classifier-free diffusion guidance.
\newblock \emph{arXiv preprint arXiv:2207.12598}, 2022.
\newblock URL \url{https://arxiv.org/abs/2207.12598}.

\bibitem[Hui et~al.(2024)Hui, Yang, Cui, et~al.]{hui2024qwen25coder}
Binyuan Hui, Jian Yang, Zeyu Cui, et~al.
\newblock Qwen2.5-coder technical report.
\newblock \emph{arXiv preprint arXiv:2409.12186}, 2024.
\newblock URL \url{https://arxiv.org/abs/2409.12186}.

\bibitem[Jayaraman et~al.(2022)Jayaraman, Lambourne, Desai, Willis, Sanghi, and
  Morris]{jayaraman2022solidgen}
P.~Jayaraman, J.~Lambourne, Nishkrit Desai, Karl D.~D. Willis, Aditya Sanghi,
  and Nigel Morris.
\newblock Solidgen: An autoregressive model for direct b-rep synthesis.
\newblock \emph{Trans. Mach. Learn. Res.}, 2022.
\newblock \doi{10.48550/arXiv.2203.13944}.
\newblock URL
  \url{https://www.semanticscholar.org/paper/301bf1ecb28ae1035e6dcef5ed67add79271c252}.

\bibitem[Khan et~al.(2024)Khan, Sinha, Sheikh, Stricker, Ali, and
  Afzal]{khan2024text2cad}
Mohammad~Sadil Khan, Sankalp Sinha, Talha~Uddin Sheikh, Didier Stricker,
  Sk~Aziz Ali, and Muhammad~Zeshan Afzal.
\newblock Text2cad: Generating sequential {CAD} designs from beginner-to-expert
  level text prompts.
\newblock In \emph{Advances in Neural Information Processing Systems},
  volume~37, pages 7552--7579. Curran Associates, Inc., 2024.
\newblock URL
  \url{https://proceedings.neurips.cc/paper_files/paper/2024/hash/0e5b96f97c1813bb75f6c28532c2ecc7-Abstract-Conference.html}.

\bibitem[Lam et~al.(2025)Lam, Wang, tse Huang, and Lyu]{lam2025codecrash}
Man~Ho Lam, Chaozheng Wang, Jen tse Huang, and Michael~R. Lyu.
\newblock Codecrash: Exposing llm fragility to misleading natural language in
  code reasoning.
\newblock In \emph{Neural Information Processing Systems}, 2025.
\newblock URL \url{https://arxiv.org/abs/2504.14119}.

\bibitem[Li et~al.(2025)Li, Ma, Li, Lou, Zhou, and Zhou]{li2025cadllama}
Jiahao Li, Weijian Ma, Xueyang Li, Yunzhong Lou, Guichun Zhou, and Xiangdong
  Zhou.
\newblock {CAD-Llama}: Leveraging large language models for computer-aided
  design parametric {3D} model generation.
\newblock In \emph{Proceedings of the IEEE/CVF Conference on Computer Vision
  and Pattern Recognition (CVPR)}, pages 18563--18573, 2025.
\newblock URL
  \url{https://openaccess.thecvf.com/content/CVPR2025/html/Li_CAD-Llama_Leveraging_Large_Language_Models_for_Computer-Aided_Design_Parametric_3D_CVPR_2025_paper.html}.

\bibitem[Mallis et~al.(2026)Mallis, Wang, Karadeniz, Ricci, Kacem, and
  Aouada]{mallis2026cadtests}
Dimitrios Mallis, Marco Wang, Ahmet~Serdar Karadeniz, Elisa Ricci, Anis Kacem,
  and Djamila Aouada.
\newblock Text-to-cad evaluation with cadtests.
\newblock \emph{arXiv preprint arXiv:2605.07807}, 2026.
\newblock URL \url{https://arxiv.org/abs/2605.07807}.

\bibitem[Peng et~al.(2025)Peng, Chen, Tang, and Li]{peng2025misbench}
Miao Peng, Nuo Chen, Jianheng Tang, and Jia Li.
\newblock How does misinformation affect large language model behaviors and
  preferences?
\newblock In \emph{Proceedings of the 63rd Annual Meeting of the Association
  for Computational Linguistics}, 2025.
\newblock URL \url{https://aclanthology.org/2025.acl-long.674/}.

\bibitem[Rukhovich et~al.(2025)Rukhovich, Dupont, Mallis, Cherenkova, Kacem,
  and Aouada]{rukhovich2024cadrecode}
Danila Rukhovich, Elona Dupont, Dimitrios Mallis, Kseniya Cherenkova, Anis
  Kacem, and Djamila Aouada.
\newblock {CAD-Recode}: Reverse engineering {CAD} code from point clouds.
\newblock In \emph{Proceedings of the IEEE/CVF International Conference on
  Computer Vision (ICCV)}, pages 9801--9811, 2025.
\newblock \doi{10.1109/ICCV51701.2025.00914}.
\newblock URL
  \url{https://openaccess.thecvf.com/content/ICCV2025/html/Rukhovich_CAD-Recode_Reverse_Engineering_CAD_Code_from_Point_Clouds_ICCV_2025_paper.html}.

\bibitem[Sanchez et~al.(2023)Sanchez, Fan, Spangher, Levi, Ammanamanchi, and
  Biderman]{sanchez2023cfg}
Guillaume Sanchez, Honglu Fan, Alexander Spangher, Elad Levi, Pawan~Sasanka
  Ammanamanchi, and Stella Biderman.
\newblock Stay on topic with classifier-free guidance.
\newblock \emph{arXiv preprint arXiv:2306.17806}, 2023.
\newblock URL \url{https://arxiv.org/abs/2306.17806}.

\bibitem[Tao and Agrawal(2026)]{tao2026nwcad}
Yufei Tao and Ameeta Agrawal.
\newblock No-worse context-aware decoding: Preventing neutral regression in
  context-conditioned generation.
\newblock In \emph{Findings of the Association for Computational Linguistics:
  ACL 2026}, 2026.
\newblock URL \url{https://arxiv.org/abs/2604.16686}.

\bibitem[Waheed et~al.(2025)Waheed, Wu, Ros{\'e}, and
  Ippolito]{waheed2025codeinduced}
Abdul Waheed, Zhen Wu, Carolyn Ros{\'e}, and Daphne Ippolito.
\newblock On code-induced reasoning in llms.
\newblock \emph{arXiv preprint arXiv:2509.21499}, 2025.
\newblock URL \url{https://arxiv.org/abs/2509.21499}.

\bibitem[Wang et~al.(2025{\natexlab{a}})Wang, Prasad, Stengel-Eskin, and
  Bansal]{wang2025madamrag}
Han Wang, Archiki Prasad, Elias Stengel-Eskin, and Mohit Bansal.
\newblock Retrieval-augmented generation with conflicting evidence.
\newblock In \emph{Conference on Language Modeling (COLM)}, 2025{\natexlab{a}}.
\newblock URL \url{https://arxiv.org/abs/2504.13079}.

\bibitem[Wang et~al.(2025{\natexlab{b}})Wang, Yuan, Sun, and
  Bian]{wang2025cadfusion}
Ruiyu Wang, Yu~Yuan, Shizhao Sun, and Jiang Bian.
\newblock Text-to-{CAD} generation through infusing visual feedback in large
  language models.
\newblock In \emph{Proceedings of the 42nd International Conference on Machine
  Learning}, volume 267 of \emph{Proceedings of Machine Learning Research},
  pages 65326--65345. PMLR, 2025{\natexlab{b}}.
\newblock URL \url{https://proceedings.mlr.press/v267/wang25eg.html}.

\bibitem[Wu et~al.(2021)Wu, Xiao, and Zheng]{wu2021deepcad}
Rundi Wu, Chang Xiao, and Changxi Zheng.
\newblock Deepcad: A deep generative network for computer-aided design models.
\newblock In \emph{IEEE International Conference on Computer Vision}, 2021.
\newblock \doi{10.1109/ICCV48922.2021.00670}.
\newblock URL
  \url{https://www.semanticscholar.org/paper/57d71b30f2ff68bde4a0c50322bb93a5c3358ee0}.

\bibitem[Xie and Ju(2025)]{xie2025texttocadquery}
Haoyang Xie and Feng Ju.
\newblock Text-to-cadquery: A new paradigm for cad generation with scalable
  large model capabilities.
\newblock \emph{arXiv.org}, 2025.
\newblock \doi{10.48550/arXiv.2505.06507}.
\newblock URL
  \url{https://www.semanticscholar.org/paper/5ac4c1a706d5bc49d483f5f8c63323b0fd780a82}.

\bibitem[Xu et~al.(2022)Xu, Willis, Lambourne, Cheng, Jayaraman, and
  Furukawa]{xu2022skexgen}
Xiang Xu, Karl D.~D. Willis, J.~Lambourne, Chin-Yi Cheng, P.~Jayaraman, and
  Yasutaka Furukawa.
\newblock Skexgen: Autoregressive generation of cad construction sequences with
  disentangled codebooks.
\newblock In \emph{International Conference on Machine Learning}, 2022.
\newblock \doi{10.48550/arXiv.2207.04632}.
\newblock URL
  \url{https://www.semanticscholar.org/paper/816a64e5b7247ed0b8d29d300e2cae787674e174}.

\bibitem[Xu et~al.(2023)Xu, Jayaraman, Lambourne, Willis, and
  Furukawa]{xu2023hnccad}
Xiang Xu, Pradeep~Kumar Jayaraman, Joseph~G. Lambourne, Karl D.~D. Willis, and
  Yasutaka Furukawa.
\newblock Hierarchical neural coding for controllable {CAD} model generation.
\newblock In \emph{Proceedings of the 40th International Conference on Machine
  Learning}, volume 202 of \emph{Proceedings of Machine Learning Research},
  pages 38443--38461. PMLR, 2023.
\newblock URL \url{https://proceedings.mlr.press/v202/xu23f.html}.

\bibitem[Yuan et~al.(2026)Yuan, Zhao, Molodyk, Hu, and Chen]{yuan2026procad}
Bo~Yuan, Zelin Zhao, Petr Molodyk, Bin Hu, and Yongxin Chen.
\newblock Clarify before you draw: Proactive agents for robust text-to-cad
  generation.
\newblock \emph{arXiv preprint arXiv:2602.03045}, 2026.
\newblock URL \url{https://arxiv.org/abs/2602.03045}.

\bibitem[Zhang et~al.(2024)Zhang, Sun, Wang, Cai, and Bian]{zhang2024flexcad}
Zhanwei Zhang, Shizhao Sun, Wenxiao Wang, Deng Cai, and Jiang Bian.
\newblock Flexcad: Unified and versatile controllable cad generation with
  fine-tuned large language models.
\newblock \emph{arXiv (Cornell University)}, 2024.
\newblock \doi{10.48550/arxiv.2411.05823}.
\newblock URL \url{https://doi.org/10.48550/arxiv.2411.05823}.

\bibitem[Zuo et~al.(2025)Zuo, Gan, Long, and Liu]{zuo2025cadhllm}
Zhuo Zuo, Yantao Gan, Junfeng Long, and Xianggen Liu.
\newblock Cad-hllm: Generating executable cad from text with hierarchical llm
  planning.
\newblock In \emph{Proceedings of the Asian Conference on Machine Learning
  (ACML)}, volume 304 of \emph{Proceedings of Machine Learning Research}, 2025.

\end{thebibliography}

\newpage
\appendix
\section*{Appendix}

\section{Pre-registration record and report-only probes}

\label{sec:appendix_a_pre_registration_record_and_r}

\subsection{The two freeze chains}

\label{sec:the_two_freeze_chains}

Every decision rule in this paper was frozen, with a SHA-256 hash recorded, before the data it would be applied to existed. The record has two chains, one per evaluation campaign, and each chain has two links.

\textbf{Chain 1 --- the initial evaluation.} The six-cell protocol (decision rules, tests, thresholds) was frozen before any scoring. The control protocol and its analysis script (\texttt{f4\_analysis.py}, sha256 \texttt{c506ab4e\ldots{}}) were frozen on 2026-07-07, before the control model's evaluations ran on 2026-07-08.

\textbf{Chain 2 --- the replication and the control retest.} The replication protocol, its frozen 400-program stratified sample, its per-profile draw quotas and its analysis script were frozen before the replication evaluations ran; the chain includes a blinded power check executed before any direction or $p$-value had been computed (§A.2). The Test D increment protocol, its six-job evaluation list and \texttt{testd\_analysis.py} were frozen before any of R's data on the replication sample existed. The increment changes the analysis sample and nothing else: the decision rules, the competence-gate thresholds (0.30 adherence floor in both header arms, 0.10 executability margin), the test specification (same-seed paired Wilcoxon, \texttt{pratt}, one-sided, asymptotic), the reverse-direction veto and the outcome mapping are carried over verbatim from the control protocol of Chain 1. Three quantities scale with the sample: the analysis set (400 unique deduplicated programs, requiring no within-cluster averaging), the profile set (eleven four-tag profiles, with the direction guardrail counting the ten that have $n \geq 15$), and the non-zero-pair guardrail ($60 = 0.15n$). The control was not retrained; the same three derangement-trained checkpoints were re-evaluated, and M's arms were reused from the replication evaluation, so the retest consists of six additional pure-inference jobs.

No rule, threshold or test was altered after freezing in any of the four documents. The two chains are reported separately throughout and are never pooled (§6).

\subsection{The blinded power check}

\label{sec:the_blinded_power_check}

The replication protocol pre-registered a single opportunity to enlarge the sample, conditioned on an information criterion rather than on a result. The check counts, per cell and per seed, how many paired differences are non-zero --- the quantity that actually carries a Wilcoxon test on a tie-heavy distribution --- and compares it against a threshold of $60 = 0.15n$ fixed in advance, expansion being triggered only if at least two of the three seeds fall below that threshold in either tokenization of the decision cell. It was executed by \texttt{blinded\_\discretionary{}{}{}power\_\discretionary{}{}{}check.py}, verified by AST inspection to contain no mean, direction or Wilcoxon code path, and run before any effect direction or $p$-value existed.

\begin{table}[ht]
\centering
\caption{Blinded power check, executed before any effect direction or $p$-value was computed: per-seed counts of non-zero paired differences against the pre-registered threshold of 60 (0.15$n$). No cell fell below it, so the frozen rule forbidding sample expansion after the check was not triggered.}
\label{tab:9}
\small
\begin{tabularx}{\columnwidth}{XlXX}
\toprule
\textbf{Cell} & \textbf{Tokenization} & \textbf{Non-zero counts (seeds 0/1/2)} & \textbf{Seeds below threshold} \\
\midrule
shuffled (decision cell) & token & 123 / 116 / 126 & 0 \\
shuffled (decision cell) & text & 123 / 113 / 126 & 0 \\
masked (report-only) & token & 121 / 130 / 158 & 0 \\
masked (report-only) & text & 131 / 139 / 145 & 0 \\
\bottomrule
\end{tabularx}
\end{table}

Every cell cleared the threshold by roughly a factor of two, the check returned no trigger, and under the frozen rule no expansion was permitted thereafter --- including, explicitly, an expansion motivated by an unwelcome result. For orientation, the decision cell of the initial evaluation rested on 12--15 non-zero clusters; the counts are a property of the design rather than an outcome, and the two campaigns' effect levels are not compared here (§6).

\subsection{Blinding of the control retest was asymmetric}

\label{sec:blinding_of_the_control_retest_was_asymm}

When the Test D increment was frozen, M's correct- and wrong-header arms on the replication sample had already been scored: they constitute the primary result of §4.1, and their per-seed and per-profile values had been inspected and recorded. R's arms on that sample did not exist. Because the test statistic is $\mathrm{drop}_M - \mathrm{drop}_R$ and $\mathrm{drop}_R$ was unobserved, the outcome of the test was not knowable at freezing time; the retest is therefore prospective in the sense Test D requires, but it is not prospective with respect to M, and we state the distinction rather than let the word ``pre-registered'' carry both. The increment's sha256 and its commit timestamp allow the ordering to be verified independently against the released artifacts.

\subsection{Procedural events during the campaign}

\label{sec:procedural_events_during_the_campaign}

Three events are recorded for completeness; none touched a frozen artifact or a scientific parameter.

The retest was launched once with an output directory that would have appended to already-frozen replication artifacts. It was terminated within three minutes, while the first jobs were still loading model weights, and relaunched into a separate directory. The four frozen artifacts were afterwards verified byte-identical, and no control data on the replication sample existed at that point, so the increment's prospective status was not compromised.

During the replication run the launcher's progress command was changed to report sample-level progress by counting lines in each worker's \texttt{adherence\_\discretionary{}{}{}samples.jsonl} rather than parsing the evaluation loop's buffered standard output. The change is observability only: no process was restarted, and the evaluation code, the job list, the batch size and the metric were untouched.

A chaining script used \texttt{pgrep -f geom\_\discretionary{}{}{}requirements} to detect whether geometry scoring was still running; the pattern matched the script's own command line, so the script waited on itself and the blinded power check was not triggered automatically. Geometry scoring had in fact completed normally and its outputs were complete, so no data was affected; the check was executed manually.

\subsection{Artifact-integrity verification}

\label{sec:artifact_integrity_verification}

The replication artifacts were audited mid-run, on the thirteen cells complete at that point (5,200 scored rows), and the campaign was verified again on completion (21 of 21 cells, 8,400 adherence rows, 8.71 h wall-clock, no failed or re-run job; geometry scoring of 8,400 rows in 3,305 s, 5,441 executable). The audit checked structure rather than sampling it.

\begin{table}[ht]
\centering
\caption{Pre-launch verification of the frozen artifacts and of the evaluation path.}
\label{tab:10}
\small
\begin{tabularx}{\columnwidth}{Xl}
\toprule
\textbf{Check} & \textbf{Result} \\
\midrule
400 rows per cell, indices 0--399 complete and unique, no parse failures & 5,200 / 5,200 rows across 13 cells \\
\texttt{gt\_code} byte-identical for a given index across every cell and seed & 0 mismatches \\
\texttt{sha256(gt\_code)} against the frozen list's \texttt{program\_sha256} & 400 / 400 \\
four-tag profile recomputed from \texttt{gt\_code} against the frozen profile & 400 / 400 \\
prompt reconstructible byte-exactly as header $+$ \texttt{gt\_code[:0.4]} & 5,200 / 5,200 \\
correct-header arm equals $\phi(\texttt{gt\_code})$ & 1,200 / 1,200 \\
masked arm carries no header; shuffled arm differs from its own correct header on the five-tag header & 1,200 / 1,200 and 1,600 / 1,600 \\
shuffled arm differs from its own correct header \textit{on the four scored tags} & 371 / 378 / 367 of 400 per seed \\
text and token arms share one shuffle assignment & 400 / 400 \\
generated programs syntactically valid; no empty, corrupted, prompt-leaking or double-import output & 0 / 5,200 failures \\
recorded executability bit reproducible on re-execution & 260 / 260 \\
generation diversity per cell & 274--340 distinct programs \\
\bottomrule
\end{tabularx}
\end{table}

Re-execution of the failing programs classified them as genuine model errors rather than harness breakage: 131 zero-length extrude geometry failures and 80 calls to methods that do not exist (\texttt{.workplanepos}, \texttt{.lineto}, \texttt{.workplanex}).

The retest artifacts were verified on the same pattern (six cells ${\times}$ 400 rows, indices complete and consistent across cells, zero \texttt{gt\_code} mismatches against the frozen list; 2,400 adherence rows in $\approx$1.9 h, 1,313 executable). One check is specific to Test D and is load-bearing rather than routine: M and R must receive the byte-identical injected wrong header at every (index, seed), since the interaction statistic is otherwise meaningless. This held on 1,200 / 1,200 comparisons, for the header string and for the full prompt, and it is not automatic --- the shuffle plan must be regenerated identically by two independent processes consuming the same random sequence. The analysis input was formed by concatenating the two campaigns' \texttt{geom\_scores.jsonl} files into \texttt{testd\_\discretionary{}{}{}analysis\_\discretionary{}{}{}input/\discretionary{}{}{}geom\_\discretionary{}{}{}scores.jsonl} (10,800 rows); both source files' hashes were unchanged by the concatenation, so the frozen analysis script's own hash remained valid and it required no modification.

\subsection{Disclosures recorded before the decision criteria were computed}

\label{sec:disclosures_recorded_before_the_decision}

The frozen protocol requires that anything discovered in flight be written down before any criterion is evaluated. Six items were.

\textbf{Prefix feature leakage is higher on the stratified sample than on the initial one.} On the 400-program sample, 17.8\% of programs have at least one four-tag intent feature already satisfied inside the 40\% prefix (18.0\% counting MULTI\_PART), against 8.0\% (10.0\%) on the initial sample; the mean fraction of a program's intent pre-satisfied by its prefix rises from 0.105 to 0.161. The composition shifts from $\{\mathrm{THIN}\,8, \mathrm{NGON}\,2, \mathrm{CIRCLE}\,2\}$ to $\{\mathrm{THIN}\,42, \mathrm{CIRCLE}\,34, \mathrm{NGON}\,11, \mathrm{TALL}\,9\}$. The cause is the stratified draw itself, which introduces the CIRCLE-family profiles the initial sample largely lacked (pure CIRCLE 3 $\rightarrow$ 40, CIRCLE+THIN 9 $\rightarrow$ 58, CIRCLE+TALL 0 $\rightarrow$ 15), and \texttt{.circle(} occurs early in a program. The prefix fraction is a frozen parameter, so the protocol was not changed; the leakage is symmetric, since every arm at a given index receives the identical prefix, and therefore cancels in the paired differences that carry every decision. Its declared consequence --- raising the recall floor, driving prefix-determined samples to identical values across arms and so increasing the zero-difference mass --- is what the non-zero-pair guardrail exists to detect (§A.2).

\textbf{Three mechanisms reduce the effective sample size, symmetrically across arms.} The generation extractor truncates at the first top-level \texttt{result =} assignment, while a multi-part program's 40\% prefix can already contain a complete assignment block; on 1,147 of 5,200 scored rows (22.1\%) the scored program therefore lies entirely inside the prompt and the model's own generation is discarded --- on the 72 indices where every compared arm is affected simultaneously, the paired difference is identically zero. The 17.8\% prefix leakage above is the second. Third, the 400 programs yield only 298 distinct 40\% prefixes, the largest collision class containing 41 programs sharing one input (spanning two intents), which the header-free arms cannot distinguish at all. Under the training-side regex, these together already give identical recall across the compared arms on 125 of 400 indices (31.2\%) before geometry scoring.

\textbf{\texttt{gen\_raw} does not store the raw completion.} The field records the result of one pass of the extractor, not the model's untrimmed output (\texttt{gen\_raw} equals the scored program on 4,937 of 5,200 rows, the remainder differing only by a trailing newline). Re-scoring the extracted programs offline under a different metric is therefore possible, and is exactly what this paper's geometry scoring does; redoing the extraction offline, or performing EOS and truncation forensics after the fact, is not. This limitation is inherited from the initial evaluation rather than introduced here.

\textbf{The \texttt{code\_sha256} field embedded in the frozen sample list is stale.} It is a snapshot taken at the sampling moment on 2026-08-05, and two of the recorded files were changed on 2026-08-06 by the parallelism and resume fixes. The authoritative record of the code actually used at launch is \texttt{replication\_\discretionary{}{}{}frozen/\discretionary{}{}{}sha256s.txt}. The frozen list itself was not edited, since editing it would invalidate its own hash.

\textbf{The regex extractor counts commented-out geometry.} A generated line such as \texttt{\# .lineTo(...)} participates in the $\geq 5$-segment test for NGON. This can only inflate the \textit{produced} side, never the intended side, and it applies identically to every arm.

\textbf{The \texttt{header\_injected} log field is written in sorted order but injected in vocabulary order.} On 1,771 of the 4,000 header-bearing rows the two orders differ, so that log field alone cannot reconstruct the prompt byte-for-byte. This affects log forensics only: the prompts themselves are stored in full, and the reconstruction check in §A.5 is run against them.

\subsection{Report-only probes}

\label{sec:report_only_probes}

\textbf{Content-blindness probe (M1).} On the replication sample R's correct-versus-wrong adherence difference (execution-inclusive) is $-0.006$ / $+0.016$ / $-0.003$ per seed, against the frozen protocol's expectation of $\approx 0$, with signs inconsistent across seeds as immunity predicts and no seed showing an execution artifact. On the initial sample the same probe gave $-0.007$ / $-0.007$ / $-0.059$, where the seed-2 value (nominal $p = 0.013$) was an execution artifact of exactly four zero-filled samples (held-out indices 56, 76, 77, 97: R-correct failed to execute and was scored 0 while R-wrong executed) and collapsed to $-0.012$ (1 non-zero pair, n.s.) on the executable intersection; that artifact does not recur on the larger sample. M1 feeds neither the competence gate nor Test D in either run, and the ruling is unchanged by it.

\textbf{Presence probe.} On the initial sample R's correct-versus-masked difference is $0.000$ / $+0.066$ / $-0.053$. This is a \textit{presence} control and, per the frozen protocol, is not citable as content evidence. It was not repeated on the replication sample: the retest ran R's correct and wrong arms only, so no masked-arm data for R exists there.

\subsection{What the released hash manifest covers}

\label{sec:what_the_released_hash_manifest_covers}

The manifest covers the analysis scripts of all four protocols; both frozen held-out lists --- the initial 100-entry set (50 unique programs) with its 76-entry / 38-program four-feature analysis subset and six profile definitions, and the replication's 400 deduplicated programs with their eleven profile definitions and per-profile stratification quotas --- and all evaluation-job lists (42 six-cell jobs; 18 initial control jobs; 21 replication jobs; 6 control-retest jobs). The concatenated analysis input from which the retest verdict was computed is released with its own hash, since neither results directory reproduces the retest table on its own.

\section{Robustness of Test D, and instrument checks}

\label{sec:appendix_b_robustness_of_test_d_and_inst}

Every check in this appendix is computed on the 400-program replication sample from \texttt{testd\_\discretionary{}{}{}analysis\_\discretionary{}{}{}input/\discretionary{}{}{}geom\_\discretionary{}{}{}scores.jsonl}, reusing the frozen test mechanics so that each per-seed statistic is identical in specification to the one reported in §4.2. The checks are report-only: the pre-registered decision rules are the competence gate, the interaction test and its three guardrails, all specified in §3.5 and frozen as described in §3.6, and nothing below can overturn or rescue them. Values from the initial evaluation appear only where they are labelled as such, and the two campaigns' effect levels are never placed side by side on a common scale.

\textbf{Zero-handling robustness.} The per-seed Test D interaction under five treatments of the tie-heavy difference distribution. Test D remains significant on every seed under all five, and no treatment moves any seed within two orders of magnitude of $\alpha$; the frozen choice (\texttt{pratt}) retains the tie-heavy zero mass rather than discarding it, and is neither uniformly the largest nor uniformly the smallest $p$ in a row, so the ruling does not turn on which treatment is used.

\begin{table}[ht]
\centering
\caption{Zero-handling robustness: per-seed Test D interaction $p$-values under five treatments of the tie-heavy difference distribution.}
\label{tab:11}
\small
\begin{tabularx}{\columnwidth}{lXXXXl}
\toprule
\textbf{Seed} & \textbf{pratt (frozen)} & \textbf{drop-zero asymptotic} & \textbf{drop-zero exact} & \textbf{sign-flip permutation (200k)} & \textbf{sign test} \\
\midrule
0 & $5.9{\times}10^{-7}$ & $1.4{\times}10^{-5}$ & $2.0{\times}10^{-5}$ & <$5{\times}10^{-6}$ & $3.8{\times}10^{-7}$ \\
1 & $5.6{\times}10^{-14}$ & $1.9{\times}10^{-13}$ & $8.6{\times}10^{-14}$ & <$5{\times}10^{-6}$ & $7.4{\times}10^{-14}$ \\
2 & $7.5{\times}10^{-17}$ & $4.7{\times}10^{-17}$ & <$10^{-15}$ & <$5{\times}10^{-6}$ & $9.0{\times}10^{-17}$ \\
\bottomrule
\end{tabularx}
\end{table}

\textit{(The permutation column is bounded below by the 200,000-draw resolution, $1/200{,}001 \approx 5{\times}10^{-6}$; no permuted mean reached the observed value on any seed. The seed-2 drop-zero exact $p$ underflows double precision and is reported as a bound.)}

\textbf{Diff-mass decomposition.} A tie-heavy distribution invites the objection that the zero mass conceals two opposing effects that cancel. It does not: the zeros are samples on which neither model moves, and the non-zero mass is dominated by the cell the causal account predicts --- M perturbed by the wrong header, R unmoved by it.

\begin{table}[ht]
\centering
\caption{Diff-mass decomposition of the 400 per-sample Test D pairs, per seed.}
\label{tab:12}
\small
\begin{tabularx}{\columnwidth}{llXXXXX}
\toprule
\textbf{Seed} & \textbf{Both flat} & \textbf{M moves, R flat} & \textbf{R moves, M flat} & \textbf{Both move, cancel} & \textbf{Interaction > 0} & \textbf{Interaction < 0} \\
\midrule
0 & 242 & 142 & 4 & 2 & 109 & 47 \\
1 & 222 & 138 & 21 & 7 & 133 & 38 \\
2 & 240 & 147 & 8 & 1 & 130 & 29 \\
\bottomrule
\end{tabularx}
\end{table}

\textit{(Pairs where both models move without exactly cancelling --- 10 / 12 / 4 per seed --- are classified only by their interaction sign, so the sign columns, not the first four, total the 156 / 171 / 159 non-zero pairs.)} The cell in which M moves and R does not outnumbers its mirror image by roughly 36-fold, 7-fold and 18-fold across seeds.

\textbf{Leave-one-profile-out, and leave-one-feature-out.} Test D re-run with each intent profile's samples removed, and then with each intent \textit{feature}'s profiles removed --- the stricter form of the check, since a feature removal deletes every profile containing that feature. All eleven single-profile removals leave the dissociation at 3/3 direction and 3/3 significance.

\begin{table}[ht]
\centering
\caption{Leave-one-profile-out and leave-one-feature-out re-runs of Test D on the replication sample.}
\label{tab:13}
\small
\begin{tabularx}{\columnwidth}{llXXX}
\toprule
\textbf{Removed} & \textbf{$n$ removed} & \textbf{Interaction mean (s0/s1/s2)} & \textbf{$p$ (s0/s1/s2)} & \textbf{Direction / significant} \\
\midrule
--- (full sample, reference) & 0 & +0.130 / +0.225 / +0.233 & $5.9{\times}10^{-7}$ / $5.6{\times}10^{-14}$ / $7.5{\times}10^{-17}$ & 3/3 \\
CIRCLE & 40 & +0.122 / +0.231 / +0.236 & $6.7{\times}10^{-6}$ / $1.1{\times}10^{-12}$ / $2.7{\times}10^{-15}$ & 3/3 \\
CIRCLE+NGON & 15 & +0.126 / +0.229 / +0.235 & $3.3{\times}10^{-6}$ / $1.9{\times}10^{-13}$ / $5.4{\times}10^{-16}$ & 3/3 \\
CIRCLE+\discretionary{}{}{}NGON+\discretionary{}{}{}TALL & 6 & +0.126 / +0.228 / +0.233 & $1.4{\times}10^{-6}$ / $7.7{\times}10^{-14}$ / $2.4{\times}10^{-16}$ & 3/3 \\
CIRCLE+\discretionary{}{}{}NGON+\discretionary{}{}{}THIN & 15 & +0.131 / +0.227 / +0.232 & $1.2{\times}10^{-6}$ / $1.8{\times}10^{-13}$ / $5.2{\times}10^{-16}$ & 3/3 \\
CIRCLE+TALL & 15 & +0.128 / +0.229 / +0.236 & $1.6{\times}10^{-6}$ / $1.7{\times}10^{-13}$ / $2.6{\times}10^{-16}$ & 3/3 \\
CIRCLE+THIN & 58 & +0.144 / +0.249 / +0.269 & $8.5{\times}10^{-7}$ / $8.4{\times}10^{-13}$ / $3.4{\times}10^{-17}$ & 3/3 \\
NGON & 39 & +0.083 / +0.191 / +0.211 & $3.3{\times}10^{-4}$ / $3.8{\times}10^{-10}$ / $7.6{\times}10^{-14}$ & 3/3 \\
NGON+TALL & 15 & +0.118 / +0.210 / +0.219 & $7.8{\times}10^{-6}$ / $3.5{\times}10^{-12}$ / $8.4{\times}10^{-15}$ & 3/3 \\
NGON+THIN & 70 & +0.146 / +0.248 / +0.273 & $4.0{\times}10^{-7}$ / $2.0{\times}10^{-14}$ / $5.9{\times}10^{-20}$ & 3/3 \\
TALL & 15 & +0.103 / +0.195 / +0.210 & $2.8{\times}10^{-5}$ / $3.3{\times}10^{-11}$ / $2.0{\times}10^{-14}$ & 3/3 \\
THIN & 112 & +0.222 / +0.250 / +0.205 & $5.7{\times}10^{-12}$ / $8.0{\times}10^{-15}$ / $6.3{\times}10^{-11}$ & 3/3 \\
\textit{all CIRCLE-containing} & 149 & +0.127 / +0.283 / +0.297 & $4.8{\times}10^{-4}$ / $5.0{\times}10^{-10}$ / $5.3{\times}10^{-13}$ & 3/3 \\
\textit{all NGON-containing} & 160 & +0.054 / +0.196 / +0.237 & $6.0{\times}10^{-2}$ / $2.1{\times}10^{-7}$ / $3.0{\times}10^{-12}$ & 3/2 \\
\textit{all THIN-containing} & 255 & +0.390 / +0.389 / +0.354 & $6.1{\times}10^{-14}$ / $8.5{\times}10^{-15}$ / $1.1{\times}10^{-13}$ & 3/3 \\
\textit{all TALL-containing} & 51 & +0.083 / +0.182 / +0.198 & $1.3{\times}10^{-3}$ / $6.7{\times}10^{-9}$ / $2.0{\times}10^{-11}$ & 3/3 \\
\bottomrule
\end{tabularx}
\end{table}

This table answers a recorded weakness of the initial evaluation. There, the dissociation leaned disproportionately on the polygonal profile: pure-NGON carried the largest per-profile interaction, and removing its 22 entries left seed 0 direction-positive but at $p = 0.32$, so the initial result satisfied its $\geq 2/3$-seed rule while depending on one intent family for one of its three seeds. That dependence does not reproduce. On 39 pure-NGON programs the removal leaves 3/3 significance, and deleting all 160 NGON-\textit{containing} programs --- 40\% of the sample, and the most aggressive removal the design permits without exhausting the polygonal geometry entirely --- still leaves 3/3 direction with 2/3 significance (seed 0 at $p = 0.060$ on 69 remaining non-zero pairs), which satisfies the pre-registered rule. We read the initial-sample row as an artifact of 22 entries rather than as a property of polygonal intent. The consequence for the paper's reading is specific: the causal claim of §4.2 does not rest on any single intent family, and in particular not on the one that dominated the initial draw.

\textbf{Profile weighting.} The pre-registered guardrail is an equal-weight mean over the ten profiles with $n \geq 15$, and it holds at $+0.283$, with every counted profile individually positive and the eleventh, uncounted profile positive as well. Two descriptive alternatives bracket it: equal weight over all eleven profiles gives $+0.276$, and sample-size weighting --- which is the pooled per-sample mean, and therefore the quantity the significance test itself operates on --- gives $+0.196$. The gap between the two weightings is the heterogeneity of Appendix C rather than a disagreement about direction: the large profiles carry smaller interactions, so equal weighting lifts the mean, and no weighting brings it near zero. Test D itself is a per-sample pooled test and is independent of any profile weighting.

\textbf{The executable-only view, and the objection it addresses.} Removing the execution-inclusive zero fill enlarges the interaction rather than dissolving it. On the four-arm executable intersection --- the pre-registered guardrail, computed only where both of M's arms and both of R's arms all produce runnable geometry --- the interaction is $+0.197$ ($N = 114$), $+0.316$ ($N = 191$) and $+0.278$ ($N = 162$) per seed, pooling to $+0.274$ over $N = 467$; averaging each arm over its own executable outputs instead gives $+0.306$ / $+0.330$ / $+0.320$. The zero fill is thus the conservative direction, not the source of the effect.

This view carries more weight than a robustness check normally would, because it speaks to the one substantive alternative reading of R's flat correct-to-wrong difference. R is an impaired control (§6): its adherence is $0.267$ / $0.354$ / $0.407$ per seed under a correct header and $0.273$ / $0.338$ / $0.410$ under a wrong one, so seed 0 sits below the pre-registered floor of 0.30 on both arms and the competence gate carries only through its $\geq 2/3$-seed rule; R is weaker than the roughly baseline-level control the frozen protocol anticipated, and a model with less adherence has, trivially, less adherence to lose. If $\mathrm{drop}_R \approx 0$ were headroom rather than immunity, R should sit near a floor on the outputs it does produce. It does not. On executable outputs the arm means are $0.673$ for the unconditioned baseline, $0.628$ and $0.621$ for R's correct and wrong arms, and $0.836$ and $0.510$ for M's. R is detectably below the baseline (Mann--Whitney two-sided, pooled over R's two arms, $p = 0.015$; $n = 655$ and $658$ against $724$) --- on the initial sample the same comparison showed no detectable difference --- but it is nowhere near zero, and it holds a range of roughly 0.62 of adherence that a wrong header could have taken from it. It loses $0.007$ of that range. M, starting from $0.836$, loses $0.326$. The objection is therefore bounded rather than dismissed: R has adherence to lose and does not lose it, on the same programs, in the same arm, under the same injected headers on which M loses a third of its own.

\textbf{Header identity is matched by construction.} The frozen retest script asserts, before any statistic is computed, that M and R receive the byte-identical injected wrong header at every (index, seed). Header content and therefore header length are matched exactly across the interaction's two arms as a checked property of the run rather than as an argument about the design. The correlation of $r \approx 0.26$ between M's per-sample drop and the header-length change reported on the initial sample was not recomputed here.

\textbf{Metric gameability.} The primary metric is not inflatable by emitting every feature; the measured feature counts per executable output are given in §3.4.

\textbf{What the two detectors read off the same reference programs.} Executing all 400 ground-truth programs and applying $\psi$ to the resulting solids gives the per-tag comparison against $\phi$ below. The disagreement is dominated by programs that do not execute at all, on which $\psi$ can read nothing; on the 271 that do, the two agree on 84.5--97.0\% of programs for the four scored tags and on 77.5\% for MULTI\_PART. Where they disagree on an executing program it is almost always $\phi$ reading a feature off the text that the solid does not realize, the exception being NGON and TALL, which $\psi$ finds on 17 and 13 programs where $\phi$ does not --- $\phi$ measures the aspect ratio from the first extrude and the largest planar extent in the text, $\psi$ from the executed bounding box.

\begin{table}[ht]
\centering
\caption{The training-side extractor $\phi$ and the geometric assertions $\psi$ compared per tag on the 400 ground-truth programs themselves. Disagreement is dominated by the 129 reference programs that do not execute, on which $\psi$ can read nothing; the last column restricts to the 271 that do.}
\label{tab:14}
\small
\begin{tabularx}{\columnwidth}{lrrrrrXX}
\toprule
\textbf{Tag} & \textbf{$\phi=1$} & \textbf{$\psi=1$} & \textbf{both} & \textbf{$\phi$ only} & \textbf{$\psi$ only} & \textbf{agreement, all 400} & \textbf{agreement, 271 executable} \\
\midrule
CIRCLE & 149 & 70 & 70 & 79 & 0 & 80.2\% & 97.0\% \\
NGON & 160 & 111 & 94 & 66 & 17 & 79.2\% & 84.5\% \\
THIN & 255 & 134 & 132 & 123 & 2 & 68.8\% & 87.5\% \\
TALL & 51 & 53 & 40 & 11 & 13 & 94.0\% & 94.8\% \\
MULTI\_PART & 202 & 34 & 34 & 168 & 0 & 58.0\% & 77.5\% \\
\bottomrule
\end{tabularx}
\end{table}

\textbf{Exclusion sensitivity, and how much of it is composition.} Three exclusions were requested of this work: the 62 zero-depth ground-truth programs, the per-seed 29/22/33 wrong headers that are null treatments in the four scored tags, and the 79 programs whose four-tag target changes under first-\texttt{result} truncation. Applied together they enlarge the effect, but a content-blind placebo filter matched on the baseline-recall distribution reproduces most of that enlargement, which is the diagnostic that separates a substantive gain from a compositional one. All figures in the table below are at the program unit, which §4.1 demotes as anti-conservative; they are reported at that unit because the exclusions are defined on ground-truth code and are most directly interpretable there, and the cluster-unit consequence is given separately for the one cell it decides.

\begin{table}[ht]
\centering
\caption{Exclusion sensitivity of the text wrong-versus-baseline effect, per seed, beside the two diagnostics that separate a substantive gain from a compositional one: direct standardization of the retained samples to the full sample's baseline-recall composition, and a content-blind placebo filter removing the same number of samples matched on that distribution.}
\label{tab:15}
\small
\begin{tabularx}{\columnwidth}{lrXXX}
\toprule
\textbf{Seed} & \textbf{unexcluded} & \textbf{all three exclusions} & \textbf{standardized to the full sample's baseline composition} & \textbf{content-blind placebo filter} \\
\midrule
0 & $-0.0717$ & $-0.1250$ & $-0.0902$ & $-0.1112$ (74\% of the gain) \\
1 & $-0.0563$ & $-0.0859$ & $-0.0575$ & $-0.0810$ (83\% of the gain) \\
2 & $-0.1150$ & $-0.1571$ & $-0.1228$ & $-0.1472$ (76\% of the gain) \\
\bottomrule
\end{tabularx}
\end{table}

The one cell that fails at the cluster level (text, seed 1) runs across the eight subsets of these three exclusions as follows: none $p = 0.098$, header-only $0.061$, truncation-only $0.101$, truncation+header $0.060$, zero-depth $0.027$, zero-depth+header $0.018$, zero-depth+truncation $0.031$, all three $0.020$. Only the four specifications containing the zero-depth exclusion place it below 0.05, and the truncation exclusion on its own moves it the wrong way. We report the cell as failing.

\textbf{Batch size on the primary metric.} §3.6 records that greedy decoding is not batch-invariant in this implementation and that the batch size is frozen at 8 for that reason. The magnitude that matters is the effect on the score rather than on the text, measured before launch on 64 frozen samples under a matched condition (text header, correct content, 40\% prefix, one checkpoint).

\begin{table}[ht]
\centering
\caption{Effect of the evaluation batch size on the primary metric rather than on the generated text, measured before launch on 64 frozen samples under a matched condition (one checkpoint, text header, correct content, 40\% prefix).}
\label{tab:16}
\small
\begin{tabularx}{\columnwidth}{XXXX}
\toprule
\textbf{Comparison} & \textbf{generations byte-identical} & \textbf{primary metric identical per sample} & \textbf{shift in the arm mean} \\
\midrule
batch 8, re-run at batch 8 & 64 / 64 & 64 / 64 & $0.0000$ \\
batch 8 against batch 16 & 56 / 64 & 62 / 64 & $-0.0234$ \\
batch 8 against batch 32 & 50 / 64 & 62 / 64 & $+0.0078$ \\
\bottomrule
\end{tabularx}
\end{table}

A 22\% divergence in the text moves the score on 3\% of samples, and the shift reverses sign between the two comparisons, so it behaves as noise rather than as a systematic bias. Averaging over 400 rather than 64 samples would shrink a purely random component to roughly 0.009, about a fifth of the smallest cell this paper reports (the cluster-level text seed-1 effect of $-0.041$), but that scaling is valid only for the random part. The measurement is single-arm, so it bounds the noise and says nothing about an interaction between padding width and treatment --- which is the actual concern, since header length differs by arm and therefore so does the left-pad width. Only a batch-size-1 re-run settles that, and it is listed among the follow-ups.

\section{Per-profile detail}

\label{sec:appendix_c_per_profile_detail}

Everything in this appendix is descriptive. The frozen analysis defines no per-profile test, $p$-value or power, and the two pre-registered guardrails act on the mean across profiles, never on an individual row. No claim in the paper is attached to an individual intent family, in either direction.

\textbf{Per-profile detail on the replication sample.} Text headers, 40\% prefix, seed-averaged four-feature execution-inclusive recall over the 400 deduplicated held-out programs. The table reports both contrasts the paper uses together with the five arm levels they are computed from, because the arm levels are what make the two contrasts legible: \texttt{wrong - baseline} is measured against each profile's \textit{unconditioned} floor, whereas the interaction is measured against the \textit{correct-header} arm, so a profile with a very low unconditioned baseline can show a positive \texttt{wrong - baseline} and a large positive interaction at once. On this sample no profile is floored in the sense the initial table reported --- the smallest unconditioned baseline is 0.133 --- and every profile's interaction is positive.

\begin{table}[ht]
\centering
\caption{Per-profile detail on the replication sample. Text headers, 40\% prefix, seed-averaged four-feature execution-inclusive recall; both contrasts are shown beside the arm levels they are computed from. Descriptive: the pre-registered guardrails act on the column means, never on a row.}
\label{tab:17}
\small
\adjustbox{max width=\columnwidth}{%
\large\setlength{\tabcolsep}{2pt}
\begin{tabular}{lrrrrrrrr}
\toprule
\textbf{Profile} & \textbf{$n$} & \textbf{baseline} & \textbf{M correct} & \textbf{M wrong} & \textbf{R correct} & \textbf{R wrong} & \textbf{wrong - baseline} & \textbf{interaction} \\
\midrule
THIN & 112 & 0.592 & 0.568 & 0.440 & 0.446 & 0.438 & -0.152 & +0.119 \\
NGON+THIN & 70 & 0.438 & 0.400 & 0.305 & 0.357 & 0.331 & -0.133 & +0.069 \\
CIRCLE+THIN & 58 & 0.279 & 0.284 & 0.233 & 0.239 & 0.236 & -0.046 & +0.049 \\
CIRCLE & 40 & 0.475 & 0.525 & 0.342 & 0.475 & 0.483 & -0.133 & +0.192 \\
NGON & 39 & 0.179 & 0.786 & 0.291 & 0.188 & 0.205 & +0.111 & +0.513 \\
CIRCLE+NGON & 15 & 0.133 & 0.367 & 0.189 & 0.122 & 0.122 & +0.056 & +0.178 \\
CIRCLE+\discretionary{}{}{}NGON+\discretionary{}{}{}THIN & 15 & 0.289 & 0.474 & 0.348 & 0.222 & 0.267 & +0.059 & +0.170 \\
CIRCLE+TALL & 15 & 0.489 & 0.533 & 0.389 & 0.511 & 0.511 & -0.100 & +0.144 \\
NGON+TALL & 15 & 0.311 & 0.744 & 0.233 & 0.289 & 0.311 & -0.078 & +0.533 \\
TALL & 15 & 0.156 & 0.978 & 0.111 & 0.133 & 0.133 & -0.044 & +0.867 \\
CIRCLE+\discretionary{}{}{}NGON+\discretionary{}{}{}TALL & 6 & 0.444 & 0.519 & 0.315 & 0.463 & 0.463 & -0.130 & +0.204 \\
\textbf{guardrail mean ($n \geq 15$)} &  &  &  &  &  &  & \textbf{-0.046} & \textbf{+0.283} \\
\bottomrule
\end{tabular}
}
\end{table}

\texttt{CIRCLE+NGON+TALL} exhausts its pool at six programs; per the frozen protocol it is reported but excluded from both guardrail means.

\textbf{Reading the two contrast columns.} Eight of the eleven \texttt{wrong - baseline} point estimates are negative and three positive; all eleven interaction point estimates are positive. The one large positive \texttt{wrong - baseline} estimate belongs to pure NGON, at $+0.111$. That profile's unconditioned baseline is 0.179, the third-lowest in the table --- this checkpoint can barely produce a polygonal loop without a header --- so a wrong header still leaves generation above that floor even though the correct-to-wrong fall on the same programs, $0.786 \rightarrow 0.291$, is among the largest recorded. Pure TALL is the same arithmetic at its extreme: baseline 0.156, correct 0.978, wrong 0.111, giving a \texttt{wrong - baseline} of only $-0.044$ alongside the table's largest interaction, $+0.867$. The permitted reading is that the pure-NGON point estimate runs in the opposite direction to the aggregate; nothing in this design tests whether it does so reliably, and these are point estimates on 39 and 15 programs offered as an account of the arithmetic, not as per-family findings.

\textbf{Per-profile arm levels under the token header and under masking.} The token-side and masked-side levels are given separately because they are what makes the aggregate token result of §4.4 legible. The three largest profiles, all of them THIN-bearing, account for 240 of the 400 programs, and on all three the token correct-header arm sits at or below the unconditioned baseline (THIN $0.321$ against $0.592$; NGON+THIN $0.190$ against $0.438$; CIRCLE+THIN $0.213$ against $0.279$), while the large positive token gains are concentrated in the small, low-baseline profiles (TALL $0.933$ against $0.156$; NGON $0.667$ against $0.179$). The aggregate token correct-header deficit is the sum of those two facts on this sample's profile composition, not a uniform failure of the token form.

\begin{table}[ht]
\centering
\caption{Per-profile arm levels under the token header and under masking (replication sample, 40\% prefix, seed-averaged; descriptive).}
\label{tab:18}
\small
\adjustbox{max width=\columnwidth}{%
\normalsize\setlength{\tabcolsep}{4pt}
\begin{tabular}{lrrrrrr}
\toprule
\textbf{Profile} & \textbf{$n$} & \textbf{baseline} & \textbf{token correct} & \textbf{token wrong} & \textbf{token masked} & \textbf{text masked} \\
\midrule
THIN & 112 & 0.592 & 0.321 & 0.327 & 0.220 & 0.202 \\
NGON+THIN & 70 & 0.438 & 0.190 & 0.229 & 0.214 & 0.179 \\
CIRCLE+THIN & 58 & 0.279 & 0.213 & 0.201 & 0.190 & 0.190 \\
CIRCLE & 40 & 0.475 & 0.500 & 0.317 & 0.433 & 0.483 \\
NGON & 39 & 0.179 & 0.667 & 0.162 & 0.188 & 0.222 \\
CIRCLE+NGON & 15 & 0.133 & 0.322 & 0.178 & 0.256 & 0.256 \\
CIRCLE+\discretionary{}{}{}NGON+\discretionary{}{}{}THIN & 15 & 0.289 & 0.259 & 0.267 & 0.326 & 0.407 \\
CIRCLE+TALL & 15 & 0.489 & 0.522 & 0.378 & 0.422 & 0.422 \\
NGON+TALL & 15 & 0.311 & 0.622 & 0.167 & 0.156 & 0.222 \\
TALL & 15 & 0.156 & 0.933 & 0.111 & 0.044 & 0.067 \\
CIRCLE+\discretionary{}{}{}NGON+\discretionary{}{}{}TALL & 6 & 0.444 & 0.444 & 0.278 & 0.315 & 0.407 \\
\bottomrule
\end{tabular}
}
\end{table}

\textbf{Why the earlier scope restriction does not survive.} Earlier versions of this work restricted the wrong-versus-none claim to polygonal and thin intents, on the ground that pure CIRCLE and TALL were floored at baseline. That restriction was a property of the initial sample rather than of the model. The initial 76-entry analysis subset covers 38 unique programs across six profiles --- three of them pure CIRCLE, two TALL --- with no programs at all in five of the eleven profiles the clean held-out pool contains (CIRCLE+NGON, CIRCLE+NGON+TALL, CIRCLE+NGON+THIN, CIRCLE+TALL, NGON+TALL). On three and two programs, a seed-averaged 0.00 in two arms is indistinguishable from an empty cell. With fifteen or more programs per profile, the replication finds pure CIRCLE at an unconditioned baseline of 0.475 --- second only to pure THIN among the single-feature profiles --- and TALL at 0.156 with a correct-header arm of 0.978. Neither is at a floor, and pure NGON, the profile the earlier restriction relied on most, sits at an unconditioned baseline of 0.179 --- below pure CIRCLE's 0.475, and the second-lowest of the four single-feature profiles. The replacement for the restriction is disclosure of the heterogeneity (§4.3) rather than a narrower scope drawn by the same method.

\textbf{Per-profile means, initial sample (six profiles; 76 scored entries over 38 unique programs; text headers, 40\% prefix, seed-averaged).} Retained for completeness and for comparison against the released initial-run artifacts. These values must not be read on a common scale with the replication table above --- the two evaluations differ in profile composition, by construction, and in generation-truncation regime (§6) --- and they are superseded for every scoping purpose.

\begin{table}[ht]
\centering
\caption{Per-profile means on the \textit{initial} sample --- six profiles, 76 scored entries over 38 unique programs, text headers, 40\% prefix, seed-averaged. Reported for continuity with the earlier evaluation; never pooled with the replication sample or placed on a common scale with it.}
\label{tab:19}
\small
\adjustbox{max width=\columnwidth}{%
\begin{tabular}{lrrrrrrr}
\toprule
\textbf{Profile} & \textbf{$n$} & \textbf{baseline} & \textbf{correct} & \textbf{wrong} & \textbf{drop$_M$} & \textbf{drop$_R$} & \textbf{interaction} \\
\midrule
CIRCLE & 6 & 0.000 & 0.000 & 0.000 & 0.000 & 0.000 & 0.000 \\
CIRCLE+THIN & 18 & 0.241 & 0.185 & 0.213 & -0.028 & -0.037 & 0.009 \\
NGON & 22 & 0.515 & 0.894 & 0.424 & 0.470 & -0.076 & 0.545 \\
NGON+THIN & 8 & 0.667 & 0.750 & 0.708 & 0.042 & -0.021 & 0.062 \\
TALL & 4 & 0.000 & 1.000 & 0.000 & 1.000 & 0.000 & 1.000 \\
THIN & 18 & 0.630 & 0.519 & 0.241 & 0.278 & 0.037 & 0.241 \\
\bottomrule
\end{tabular}
}
\end{table}

Non-tied pairs on the initial sample, for the wrong-versus-baseline contrast (harmful / helpful / tied of 76): token 19/0/57, 20/1/55, 19/4/53; text 12/6/58, 16/6/54, 16/4/56; positive control (correct versus baseline, token) 8/16/52, 18/12/46, 16/16/44. The corresponding counts on the replication sample are in §4.1.

\textbf{Dose-response probe, initial sample only.} The frozen control protocol pre-registered a supporting, content-graded probe on M: a \textit{corrupted} arm (one or two primary features falsified) and a \textit{random} arm (well-formed, size-matched, sample-unrelated feature sets, designated non-evidential ex ante). Arm means were correct 0.557 > corrupted 0.393 > wrong 0.305 > random 0.219, with per-seed one-sided $p$: correct > corrupted 0.091 / $3{\times}10^{-4}$ / $1{\times}10^{-3}$ (2/3); corrupted > wrong 0.074 / 0.056 / 0.038 (1/3); wrong > random 0.011 / 0.006 / 0.040 (3/3, grading header naturalness rather than content). The pre-registered monotone criterion required \textit{both} correct > corrupted and corrupted > wrong at $\geq 2/3$ seeds and was therefore \textbf{not confirmed}. These arms were measured only on the initial sample and were not re-run on the replication sample; their numbers are not carried into any sentence that also carries a 400-program figure, and the unresolved mechanism question they were meant to address remains open (§5).

\section{Material supporting no claim in the main text}

\label{sec:appendix_d_material_supporting_no_claim_}

Two components of the system and one accompanying observation are reported here rather than in the body. Neither component was evaluated on the 400-program replication sample, and neither was evaluated under the independent geometry-side metric of §3.4. No result, figure or table in the main text depends on any of them; they are recorded so that the description of what was built and run is complete, and so that a reader who encounters them in the released code knows what status they hold.

\textbf{The CADCON-CFG variant.} The variant applies classifier-free-guidance-style header dropout during training: with probability 0.5 the header is replaced by a null-content placeholder, so that the model acquires a represented unconditional mode alongside the conditioned one. This adapts classifier-free guidance from its diffusion-\textit{training} origin \cite{ho2022cfg} rather than the inference-time-only formulation of \cite{sanchez2023cfg} to autoregressive CAD fine-tuning. Everything else --- base model, adapter configuration, training corpus, optimizer schedule, seeds --- matches model M (§3.3, Table 2); the header-dropout probability is the only differing hyperparameter, and it is the only cell of Table 2 that applies to this variant alone. Neither M, the unconditioned baseline, nor the derangement control R uses header dropout.

The motivation is the one §5 returns to: if a wrong conditioning signal is routed into generation because the model has no represented alternative to conditioning on it, then giving the model such an alternative at training time is the cheapest available intervention at the point where the failure occurs. The variant has not been evaluated under the independent geometry-side metric, and the earlier, training-side regex measurements of it are not reported, because that metric shares its feature detector with the signal being conditioned on and would credit the variant on the same circular basis §3.4 removes. The variant therefore appears in this paper as a direction a mitigation could take and not as evidence that it works; measuring whether it in fact narrows the wrong-versus-none gap is left to future work, and the paper's diagnostic claims are independent of the outcome.

\textbf{The GRPO validity-reward study.} To probe whether reinforcement learning with an execution-only reward moves geometric validity independently of adherence, Group Relative Policy Optimization was applied on top of the SFT model with a binary validity reward --- 1 if and only if the program executes in CadQuery within 10 s and yields a non-empty solid --- using $G = 4$ completions per prompt, a learning rate of $5 {\times} 10^{-7}$ and 100 steps (Table 2), with an entropy coefficient of 0.01. Lightweight proxy-reward variants were also trained, to reduce the number of oracle executions per step; the observation below does not depend on the proxy details.

\textbf{The validity/adherence observation.} Under this reward, measured validity moved while measured adherence did not. Three limits keep it out of the body of the paper. The training scale is small --- 100 GRPO steps on top of an SFT checkpoint --- so nothing here bears on what a longer run would do. The validity movement itself is a within-run, small-sample descriptive result whose size and direction vary across reward variants, and we do not present it as a robust improvement in validity. And the adherence result is a null measured with an instrument of limited sensitivity to the kind of change GRPO produces: the regex adherence metric of that round registered no difference on the great majority of samples, so the finding is an absence of detectable change rather than a demonstration that adherence was not harmed. Read with those limits, the study is consistent with executability and semantic adherence behaving as separate reward channels --- the reading the code-RL literature generally takes --- but it does not establish that they are orthogonal, and no claim in this paper rests on the question either way.

\end{document}